\documentclass[12pt]{article}

\usepackage{times}
\usepackage{booktabs} 
\usepackage{graphicx}
\usepackage{subcaption}
\usepackage{mathtools}
\usepackage{amsmath}
\usepackage{url}
\usepackage{flushend}
\usepackage[textsize=tiny]{todonotes}

\newenvironment{sciabstract}{%
\begin{quote} \bf}
{\end{quote}}

\newcounter{lastnote}

\title{Kriging-Based Robotic Exploration for \\Soil Moisture Mapping Using a Cosmic-Ray Sensor}

\author
{Jaime~Pulido~Fentanes$^{1}$, Amir~Badiee$^{2}$, Tom~Duckett$^{1}$,\\
Jonathan Evans$^{3}$, Simon~Pearson$^{4}$ and Grzegorz~Cielniak$^{1}$\\
\\
\normalsize{$^{1}$Lincoln Centre for Autonomous Systems}\\
\normalsize{School of Computer Science, University of Lincoln}\\
\normalsize{Brayford Campus, LN6 7TS, Lincoln, UK}\\
\\
\normalsize{$^{2}$School of Engineering, University of Lincoln}\\
\normalsize{Brayford Campus, LN6 7TS, Lincoln, UK}\\
\\
\normalsize{$^{3}$Centre for Ecology $\&$ Hydrology}\\
\normalsize{Wallingford, Oxfordshire, OX10 8BB, UK}\\
\\
\normalsize{$^{3}$Lincoln Institute for Agri-food Technology, University of Lincoln}\\
\normalsize{Riseholme Park, LN2 2LG, Lincoln, UK}\\
\\
\normalsize{$^\ast$Email:  jpulidofentanes@lincoln.ac.uk.}
}

\date{}

\begin{document}




\maketitle

\begin{sciabstract}

Soil moisture monitoring is a fundamental process to enhance agricultural outcomes and to protect the environment. The traditional methods for measuring moisture content in soil are laborious and expensive, and therefore there is a growing interest in developing sensors and technologies which can reduce the effort and costs. In this work, we propose to use an autonomous mobile robot equipped with a state-of-the-art non-contact soil moisture sensor that builds moisture maps on the fly and automatically selects the most optimal sampling locations. The robot is guided by an autonomous exploration strategy driven by the quality of the soil moisture model which indicates areas of the field where the information is less precise. The sensor model follows the Poisson distribution and we demonstrate how to integrate such measurements into the kriging framework. We also investigate a range of different exploration strategies and assess their usefulness through a set of evaluation experiments based on real soil moisture data collected from two different fields. We demonstrate the benefits of using the adaptive measurement interval and adaptive sampling strategies for building better quality soil moisture models. The presented method is general and can be applied to other scenarios where the measured phenomena directly affects the acquisition time and needs to be spatially mapped.

\end{sciabstract}



\section{INTRODUCTION} \label{sec:intro}


Management of water resources is of considerable concern in different parts of the world, with many areas facing prolonged droughts, while others experience devastating floods. The availability of water in the soil is essential for vegetation. In an agricultural setting, crop health depends greatly on soil moisture. It is precisely for this reason that soil moisture monitoring is key to improving agricultural processes. Perhaps the most obvious advantage of technologies for obtaining high-resolution soil moisture maps is that they would enable highly efficient irrigation planning, for example, providing an accurate estimate of the quantity of water that should be put into a field and its required spatial distribution across the field. 

Soil moisture is typically assessed either by a direct but lengthy procedure involving collecting physical soil samples followed by lab measurements, or by hand-held instruments used to measure moisture indirectly through proxies such as surface tension (manometers), or changes in soil conductivity (e.g. time-domain reflectometry (TDR) \cite{noborio2001measurement}. All of these methods are very laborious, time consuming and expensive. Recent advances in sensing technology introduced a new, non-contact method for measuring soil moisture using fast neutron detectors (\cite{zreda2008measuring}). The neutrons are generated by cosmic rays and are reflected from the soil. The reflected neutron count is directly proportional to soil moisture content. Such sensors were successfully deployed at static locations covering large areas of land \cite{evans2016soil} but also as high-resolution variants with reduced field of view and increased sensitivity \cite{shron2017cosmic}.

The most common method for creating soil moisture maps is to use data that are manually collected at pre-determined locations in the field and extrapolate the expected measurements for unvisited regions using kriging or Gaussian Process Regression~\cite{matheron1963principles,williams2006gaussian}. This is a costly and laborious process,
especially in the case of soil moisture monitoring, where the methods and instruments used to take measurements across the field require a high amount of labour and post-processing. For this reason there is a growing interest in developing instruments and methodologies to help reduce the effort and costs, while improving the quality of the resulting soil moisture models. 

In this work, we propose to use an autonomous mobile robot equipped with a non-contact soil moisture sensor that builds soil moisture maps on the fly and automatically selects the most optimal sampling locations. The robot is guided by an autonomous exploration strategy driven by the quality of the soil moisture model (i.e. Kriging Variance) which indicates areas of the field where the information is less precise, improving overall model quality. The employed fast neutron counting sensors provide a special category of measurements in which the acquisition time directly depends on the intensity of the phenomenon: in our case, the sensor registers more neutrons in drier soils. We model the sensor using the Poisson distribution and use a special kriging variant for this type of measurements. As a result, the exploration strategy plans not only the optimal sampling location but also the required acquisition time at each sampling location.

The contributions of this work are as follows:
\begin{itemize}
\item application of a novel fast neutron counting sensor for robotic-assisted spatial mapping of soil moisture;
\item integration of the Poisson measurement model into the kriging estimation and exploration framework, which devises optimal spatial locations and measurement intervals, improving the resulting moisture models;
\item evaluation and validation of the proposed framework on data collected from two different field environments.
\end{itemize}

The remainder of the paper is structured as follows: Section~\ref{sec:related} presents related work in soil moisture surveying and robotic exploration, followed by Section~\ref{sec:method}, which details our approach to Poisson kriging and exploration for soil moisture mapping using a mobile robot. The experimental framework is presented in Section~\ref{sec:experimental}, followed by results and their analysis in Section~\ref{sec:results}, and final conclusions in Section~\ref{sec:conclusions}.

\section{RELATED WORK} \label{sec:related}
Robotic environmental monitoring applications have attracted a lot of attention in the last few years~\cite{dunbabin2012robots}. One of the advantages of using robots for environmental modelling and monitoring is that they can build models on the fly.
At the same time, many authors have discussed how to use the model itself to plan new observations for data acquisition that improve the overall model. For example, Kerry et al.~\cite{kerry2010sampling} demonstrated that kriging semivariograms are highly useful for sampling planning in precision agriculture. They proposed to use ancillary information to estimate a semivariogram and thus determine the spatial frequency of sampling based on the semivariogram parameters. 

Other researchers \cite{oliver1986combining} propose the generation of an initial set of samples to obtain a semivariogram that can be extrapolated to find new sample positions. Marchant \& Lark~\cite{marchant2007optimized} proposed an adaptive approach for optimizing reconnaissance surveys. They sampled at pre-planned positions, and calculated the probability density function of the sampling density required for the main survey in a Bayesian framework. If the requirements were not met, the number and location of observations within further phases were selected to reduce the uncertainty of the required sampling density. 
However, the effort required to survey a soil variable and simultaneously build and analyse the variance of the kriging model of the soil meant that these authors stopped short of planning the entire sampling procedure based on kriging models. 

Robots, on the other hand, are able to create and update models of their operational environments through robotic exploration. A common approach is to plan trajectories that completely cover the area assuming some prior knowledge of the environment~\cite{Bochtis2017}. Other well-known exploration techniques drive the robot towards unmapped areas of the environment. For example, greedy approaches such as \cite{koenig2001greedy} drive the robot towards the nearest location where new information can be gained. In frontier-based exploration~\cite{Yamauchi97},
the robot is driven towards the boundary between the known and unknown parts of the environment, while information driven `next-best-view' methods use reward functions to predict the utility of an unexplored location~\cite{fentanes2011algorithm}.

Many authors have proposed informative path planning (IPP) techniques for modelling physical phenomena with an unknown spatial distribution. 
These techniques address how to plan a path that maximizes sensor information~\cite{binney2013optimizing} and can be classified into two approaches: those that depend only on a priori information about the environment~\cite{hollinger2013sampling}, and adaptive sampling techniques 
that can be modified depending on the observations made~\cite{sadat2015fractal}. 
More recently, Popovic et al.~\cite{popovic2017multiresolution} proposed an adaptive informative path planning methodology to map green biomass in an agricultural setting.

Other authors have opted to use different model properties to plan robot actions. 
For example, Gao et al.~\cite{gao2018novel} propose the use of an informative sampling technique to minimize the total distance travelled by a fleet of phenotyping robots.
To do this, they model the environment using Gaussian Processes and use the model variance to plan the most informative paths for the fleet. Marchant and Ramos~\cite{marchant2014bayesian} use Gaussian Processes to plan the paths that guarantee both to observe the phenomenon of interest and improve the modelling of the same phenomenon for environmental monitoring applications such as ozone concentration across the USA. 

Other authors have chosen to use Ordinary Kriging to model in-field phenomena. Glaser et al. \cite{glaser18} use it to model soil properties perceived with a multi-spectral camera, and then use the resulting model to improve the robot localisation. Kim and Shell~\cite{kim2014distributed} proposed an augmentation of Ordinary Kriging to enable modelling of ocean current dynamics which they use for adaptive path planning in the field in ocean multi-robot scenarios. Pulido Fentanes et al.~\cite{Soil_compaction_kriging_2018} proposed a robotic exploration methodology aimed at building soil condition maps using ordinary kriging variance as a reward function for exploration. 
The current work builds upon this approach to model soil moisture measured with a novel sensor that does not follow a normal distribution.
To achieve this we combine Poisson kriging with a kriging-based exploration methodology.

\section{METHODOLOGY} \label{sec:method}

In this work, we propose a kriging-based exploration pipeline for agricultural mobile robots to facilitate efficient mapping of soil moisture. The framework combines a unique sensor model, an on-line spatial mapping component and an exploration strategy to guide the robot to the next best sampling location.

We consider a special category of measurements which are based on counting, and hence follow a Poisson distribution. An inherent property of such measurements is that their uncertainty directly depends on the length of the measurement interval. In our scenario, we use a robot-mounted soil moisture sensor (see Sec.~\ref{sec:sensor}) which counts low energy neutrons as a proxy for soil moisture. Therefore the soil moisture level will affect the amount of time the robot spends at each sampling location. For the spatial mapping we use a version of ordinary kriging which incorporates measurements following a Poisson distribution (see Sec.~\ref{sec:kriging}). We use the Kriging Variance (KV) as a reward function for the exploration strategy to plan the optimal location for each subsequent measurement. Section~\ref{sec:exp_strategies} discusses the different  exploration strategies that have been applied in this work.

The original kriging framework was presented in our previous work for mapping soil compaction~\cite{Soil_compaction_kriging_2018}. In this paper, we generalise and extend the approach to take into account measurements following a Poisson distribution. This results in exploration strategies which not only consider the optimal sampling location but also adjust the measurement duration for each reading to ensure a high-quality model.

\subsection{Soil Moisture Measurement Using a Cosmic-Ray Sensor}\label{sec:sensor}

The main sensor used in this work is based on measuring fast neutrons, which are generated by cosmic rays and reflected from the soil~\cite{zreda2008measuring}. The intensity of the reflected neutrons is affected by the hydrogen in the soil, and hence provides an indication of the soil moisture content. A neutron detector is a tube containing a gas that can convert thermal neutrons into detectable electrons by ionisation. Since the detectors are sensitive to fast neutrons only, the low energy neutrons (after colliding with the hydrogen atoms) are not counted. As a  result, a higher neutron count means more fast neutrons and corresponds to dryer soil. To improve the sensitivity of the detector to fast neutrons, a polyethylene shield is used as a moderator.

Several correction procedures need to be applied on the acquired neutron counts (which we refer to as the raw neutron count $N_{raw}$) in order to account for variations in background cosmic ray intensity, atmospheric pressure and humidity~\cite{evans2016soil}. The reference values for the corrections are established during a calibration procedure which requires reference soil moisture values to be established by direct soil moisture measurements using traditional equipment. The correction factors include:
\begin{itemize}
\item Cosmic ray intensity: 
\begin{equation}\label{eq:intensity_corr}
F_{C}=\frac{C_{0}}{C},
\end{equation}
where C is the measured neutron count rate (from the nearest monitoring station) and $C_0$ is the value measured during calibration.

\item Pressure:
\begin{equation}\label{eq:pressure}
F_{P}=\exp[\beta \left ( P-P_{0} \right )],
\end{equation}
where P is the measured barometric pressure (using a barometer), $P_0$ is an arbitrary reference value (e.g. 1010 hPA) and $\beta$ is the barometric pressure coefficient established during calibration.

\item Humidity
\begin{equation}\label{eq:humidity}
F_{Q}=1+0.00054(Q-Q_{0}),
\end{equation}
where Q is the measured humidity (derived from temperature measurements) and $Q_0$ is the average humidity during calibration.

\end{itemize}

The corrected neutron count $N_{crr}$ is obtained by multiplying the raw neutron counts by the correction factors:
\begin{equation}\label{eq:corrected neutron}
N_{crr}=N_{raw} \cdot F_{P} \cdot F_{Q} \cdot F_{C}.
\end{equation}

$N_{crr}$ can then be used to calculate Volumetric Water Content (VWC), which provides the final measure of the soil moisture. Since in this paper we mainly work with the corrected neutron counts $N_{crr}$, we refer the interested readers to \cite{evans2016soil} for further detail of the exact conversion procedure.

The summarised methodology for measuring soil moisture has been used successfully by \cite{evans2016soil}, who have established a network of soil moisture monitoring stations in the UK covering an area of 12 hectares. Although this coverage is useful for large scale soil moisture assessment, its application to individual fields in agriculture is limited. To achieve higher spatial resolutions, we have employed a high-sensitivity version of the sensor consisting of 12 neutron detectors with a bespoke polyethylene shield to limit the detection footprint of the sensor to ${\sim}10$ m. The sensor mounted on our agricultural mobile robot Thorvald can be seen in Fig.~\ref{fig:thorvald}. 

\subsection{Poisson Distribution Measurements and Sampling Regime}

Our soil moisture sensor provides the corrected neutron counts $N_{crr}$. The appropriate probabilistic model for modelling count data and events is the Poisson distribution, with parameter $\lambda$ representing the average count rate over a period of ten seconds. However the uncertainty $\sigma$ in the measurement depends directly on total neutron count over the measurement time, and is calculated as follows:

\begin{equation}\label{eq:normalised_sigma}
\sigma = \lambda \frac{\sqrt{N_{crr}}}{N_{crr}}
\end{equation}




%
%
%
%

\begin{figure}[!ht]
	\centering
	\begin{subfigure}[b]{0.48\columnwidth}
      \includegraphics[width=\columnwidth]{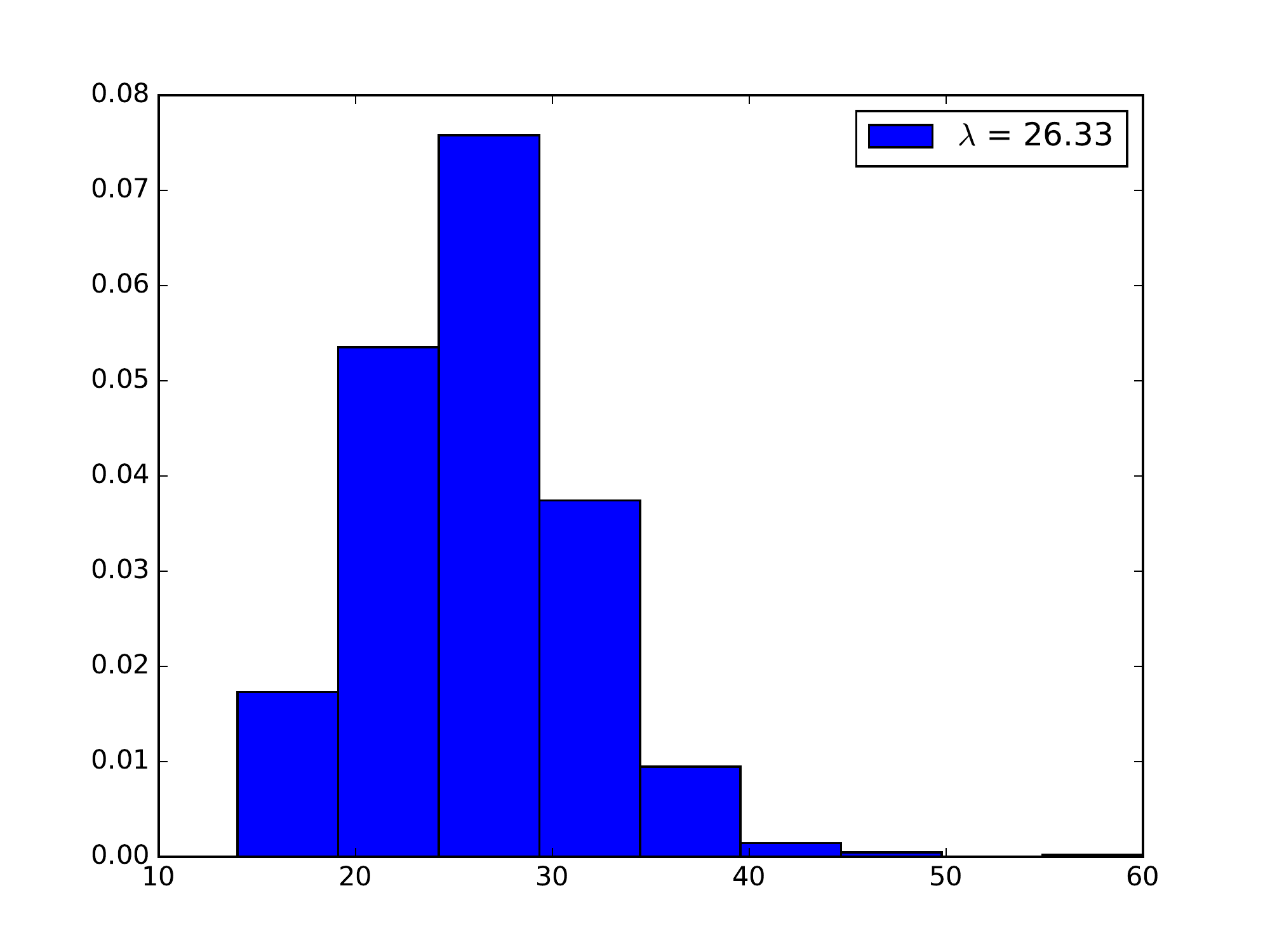}
      \caption{\label{fig:poisson_dist}}
    \end{subfigure}
	~
    \begin{subfigure}[b]{0.48\columnwidth}
      \includegraphics[width=\columnwidth]{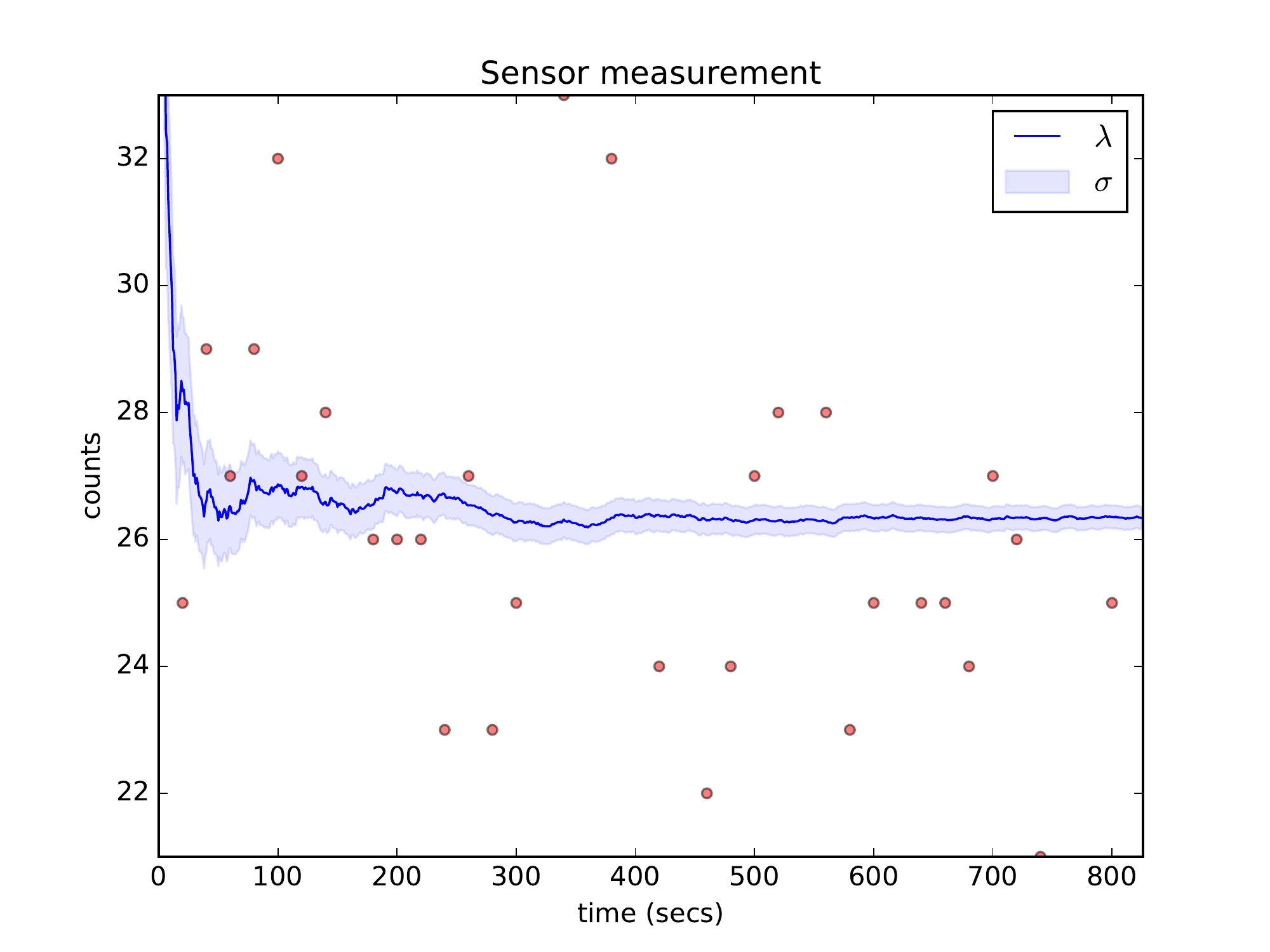}
      \caption{\label{fig:poisson_sigma}}
    \end{subfigure}
     \caption{An example measurement from the cosmic-ray sensor: a) distribution of fast neutron counts, b) evolution of the count rate and measurement uncertainty over time.}\label{fig:poisson}
\end{figure}

Figure \ref{fig:poisson} shows the histogram reading for a sensor measurement and the evolution of the $\lambda$ and $\sigma$ parameters for the same measurement over time. Figure \ref{fig:poisson_sigma} shows how the standard error and variance decrease over time, meaning that readings with longer duration achieve higher quality.

The sampling regime is the criterion used to decide how long each measurement should last. 
In this scenario, the quality of the measurement is directly correlated to the number total number of observed events ($N_{corr}$). 
For this reason, we propose to use two different methodologies: using fixed measurement intervals (FMI), in which each measurement lasts for a predetermined amount of time, or Adaptive Measurement Intervals (AMI), under which each measurement will last until a minimum level of quality is obtained.
This paper compares both regimes and analyses what happens to the exploration process with each sampling regime, and more specifically, what is their effect  on the final model quality.

\begin{figure}[!ht]
	\centering
      \includegraphics[width=0.5\columnwidth]{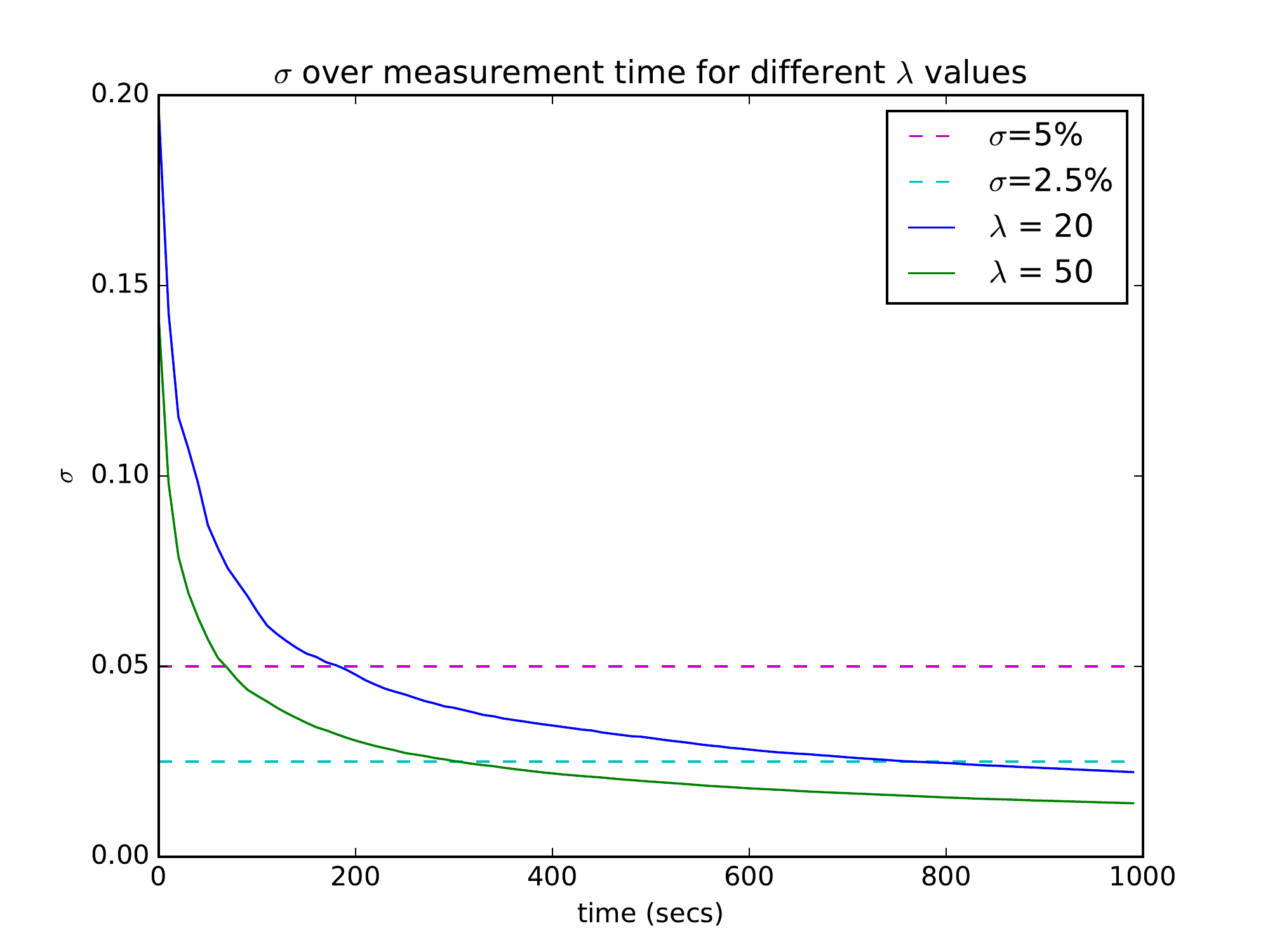}
     \caption{Measurement uncertainty $\sigma$ over time for different $\lambda$ values and sample thresholds for the AMI regime.}\label{fig:ami}
\end{figure}

The Adaptive Measurement Intervals (AMI) regime uses a threshold typically defined in terms of $\sigma_m$ (see Eq. \ref{eq:normalised_sigma}) to determine the duration of a measurement. In practice, this means that in this case the robot will stay at each location until the normalised standard error falls below a pre-determined percentage of the total amount of counts, so that the robot will stay longer in places were the count rates are lower (or the soil is wetter in this scenario) and spend less time in locations with higher count rates.

Figure \ref{fig:ami} illustrates the evolution of the normalised standard error ($\sigma_m$) over time for different rates ($\lambda$), where the dashed lines indicate thresholds that can be used for this sampling regime, the time at which the threshold lines intersect the standard error lines, represents the point at which the measurement is considered complete. This guarantees a maximum incertitude limit for each measurement which adapts to the actual neutron rate forcing the robot to stay longer at places where the rate of events is lower than usual or to leave as soon as possible in places with higher rates.

\subsection{Poisson-Kriging}\label{sec:kriging}

Ordinary kriging (OK) has proven to be an effective method for interpolating spatial data when the data's main source of error is intrinsic to the measurement technique, for example, when it depends on the precision of an instrument. 
However, when the variance of the measurement depends on the phenomenon itself, as in the case of events that can be modelled using a Poisson distribution, Ordinary Kriging does not have a way to incorporate the different variances from each data point. 

For this reason, different authors have proposed specific implementations of kriging methods that deal with data that is not normally distributed.
Monestiez et al.~\cite{monestiez2006geostatistical} presented a kriging methodology to model whale populations using data from observers on ferries and cargo ships, which can be modelled using a poisson distribution.
This approach is known as Poisson-Kriging (PK) and has since been used to model phenomena as diverse as Cancer mortality~\cite{Goovaerts2008} and gamma-ray spectral mapping~\cite{reinhart2013integrated}. For this reason, we have chosen this methodology for the current work.

PK provides an estimate $\hat{Z}(\mathbf{x}_0)$ for a variable $Z$ at unknown location $\mathbf{x}_0$ while assuming a constant unknown mean over its neighbourhood, although in this case the observations $Z(\mathbf{x}_i)$ are dependent on some underlying mean count rate and the amount of time spent at each location.
The estimate is a weighted linear combination of the available observation $z_i=Z(\mathbf{x}_i)$ and the amount of time spend at each location $t_i$ from a set of locations $\mathbf{x}_i$.
The estimator is thus described as follows:
\begin{equation} \label{eq:estimator_}
\hat{Z}(\mathbf{x}_0)=\sum_{i=1}^{n}z_i\tfrac{w_i}{t_i}, \quad i=1,\dots,n,
\end{equation}
where $\sum_{i=1}^{n}w_i=1$ to ensure unbiased estimates. To correctly estimate the values at $\mathbf{x}_0$ the weights $\mathbf{w}=[w_1,\dots,w_n]^T$ must be calculated. This can be achieved by solving the Poisson-Kriging system, which is a linear system of $n+1$ equations.

\begin{equation} \label{eq:pksystem}
\sum_{j=0}^{n}w_{ij}C_{ij}+w_i\tfrac{\hat{m}}{t_i}+\mu = C_{i\mathbf{x}_0}~~ for ~~i = 1,...,n
\end{equation}
where $\mathbf{C}_{ij}$ is the covariance of the observed values, $\mathbf{C}_{i\mathbf{x}_0}$ is the covariance at the prediction location $\mathbf{x}_0$, and $\mu$ is a Lagrange factor which ensures the optimal solution. Finally, $\hat{m}$ is estimated from the data as a weighted average of the count rates, where the weights correspond to the observation times.

Once this system is solved, the estimated values at location $\mathbf{x}_0$ can be found using Eq.~\ref{eq:estimator_}, and the associated variance of the prediction $\sigma^2$ can be calculated using the same equation as in ordinary kriging:

\begin{equation} \label{eq:kv}
\sigma^2(\mathbf{x}_0) = \sum_{i=1}^{n}w_i\mathbf{C}_{i\mathbf{x}_0} 
\end{equation}

\subsubsection{Semivariogram}

In empirical scenarios, it is possible to use a semivariogram created from the real-world data to express the relation between locations and estimate the weights for each observation. However, unlike Ordinary Kriging in this case it is necessary to account also for the observation times for each data point. For this reason PK uses a weighted variogram estimator, which takes into account the different observation times:

\begin{equation} \label{eq:semiv_est}
\hat{\gamma}(h)= \frac{1}{2N(h)}\sum_{i,j}^{n}\left ( \frac{t_it_j}{t_i+t_j}\left (\frac{z_i}{t_i}- \frac{z_j}{t_j}\right )^{2} - \hat{m} \right )I_{d_{ij} \sim h},
\end{equation}
where $h$ is the chosen distance, $\hat{m}$ is the same mean as in Eq.~\ref{eq:pksystem} and $I_{d_{ij}\sim h}$ is a gating function that takes a value of 1 when $i$ and $j$ are roughly distance $h$ apart, and 0 otherwise. $N(h)$ is a normalising factor calculated as follows:

\begin{equation} \label{eq:n_h}
N(h)=\sum_{i,j}^{n}\frac{t_it_j}{t_i+t_j}I_{d_{ij} \sim h}
\end{equation}

The semivariograms $\mathbf{\gamma}(h)$ can take multiple forms, but are generally characterised by an equation that can be parametrised.
We use the following Gaussian semivariogram model in our work: 
\begin{equation}\label{eq:gaussian_sv}
  \gamma(h )= p_0 + (p_2-p_0)(1-exp(-\frac{h^2}{p_1^2})),
\end{equation}
with the following three parameters: nugget $p_0$,  range $p_1$ and sill $p_2$~\cite{kriging_parameters_2018}.

The parameters for this equation are automatically fitted from the semivariogram of the sampled data using the “soft” $L^1$ norm minimization scheme \cite{pykrige}.

\subsection{Exploration Strategies}\label{sec:exp_strategies}

Our proposal is to use the variance of the kriging \emph{(KV)} process (see Eq.~\ref{eq:kv}) as a measurement of information gain.
The use of Kriging Variance as a reward function for robotic exploration has been previously studied in \cite{Soil_compaction_kriging_2018,kriging_parameters_2018}. 
In this work, we compare some well-known exploration strategies and how they interact with the sampling regime.
The methods to be tested can be classified into: \emph{Next-Best-View} methods and \emph{Adaptive Sampling} methods.

\subsubsection{Next-Best-View (NBV)}

These methods update the environment model every time a new sample is acquired, and then choose a new location depending on the distribution of the KV across the field. Location selection is done using one of the following strategies:

\begin{itemize}
\item \emph{Greedy}: The next sampling point is the point with the highest KV in the set of candidate locations.
\item \emph{Monte Carlo}: a set of candidate sampling locations is generated each time, and each candidate location is allocated a weight depending on its KV. The next sampling location is selected randomly, but in a way that guarantees that the probabilities are distributed according to the weight of each candidate.
\end{itemize}

\subsubsection{Adaptive Sampling}

In this category, strategies generate an initial plan that is modified depending on the KV after each model update. In this case, the robot will plan a sampling regime based on a random trajectory and a mission time horizon, which depends on the minimum expectations of measurements to be made in each case.

Every new sample taken triggers a model update, which removes sampling points based on their KV.
The targets whose KV is below the overall mean of the model are removed, and then as many new points as necessary to meet the minimum expectation of measurements in the remaining mission time are added by drawing a new waypoint from a set of candidates weighted by their KV. Finally, a new route is re-planned through the new set of points using a Traveling Salesperson (\emph{tsp}) algorithm.

\section{EXPERIMENTAL FRAMEWORK} \label{sec:experimental}

\subsection{Hardware Setup}\label{sec:thorvald}

\begin{figure}[!ht]
	\centering
	\begin{subfigure}[b]{0.48\columnwidth}
      \includegraphics[width=\columnwidth]{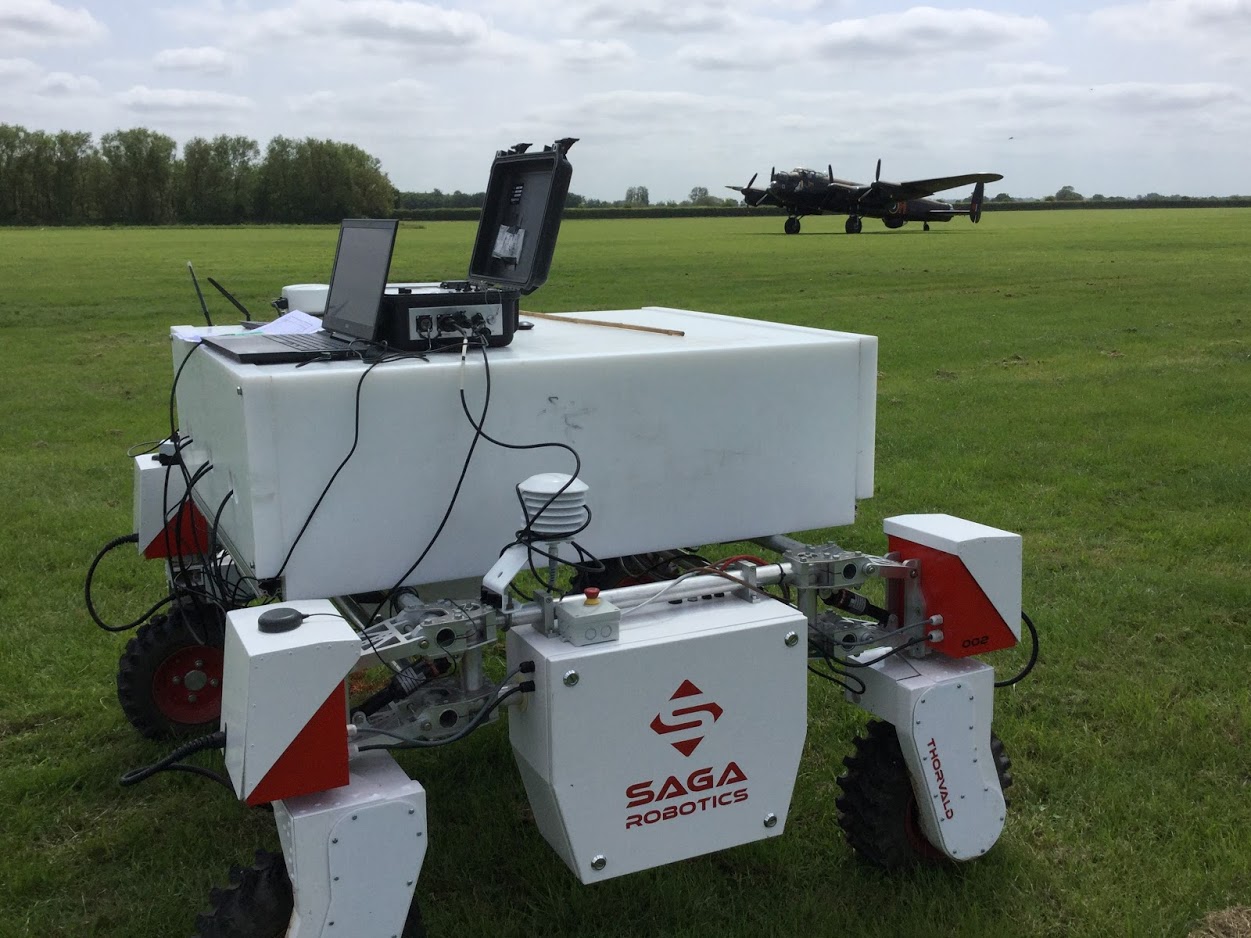}
      \caption{\label{fig:thorvald_uk}}
    \end{subfigure}
	~
    \begin{subfigure}[b]{0.48\columnwidth}
      \includegraphics[width=\columnwidth]{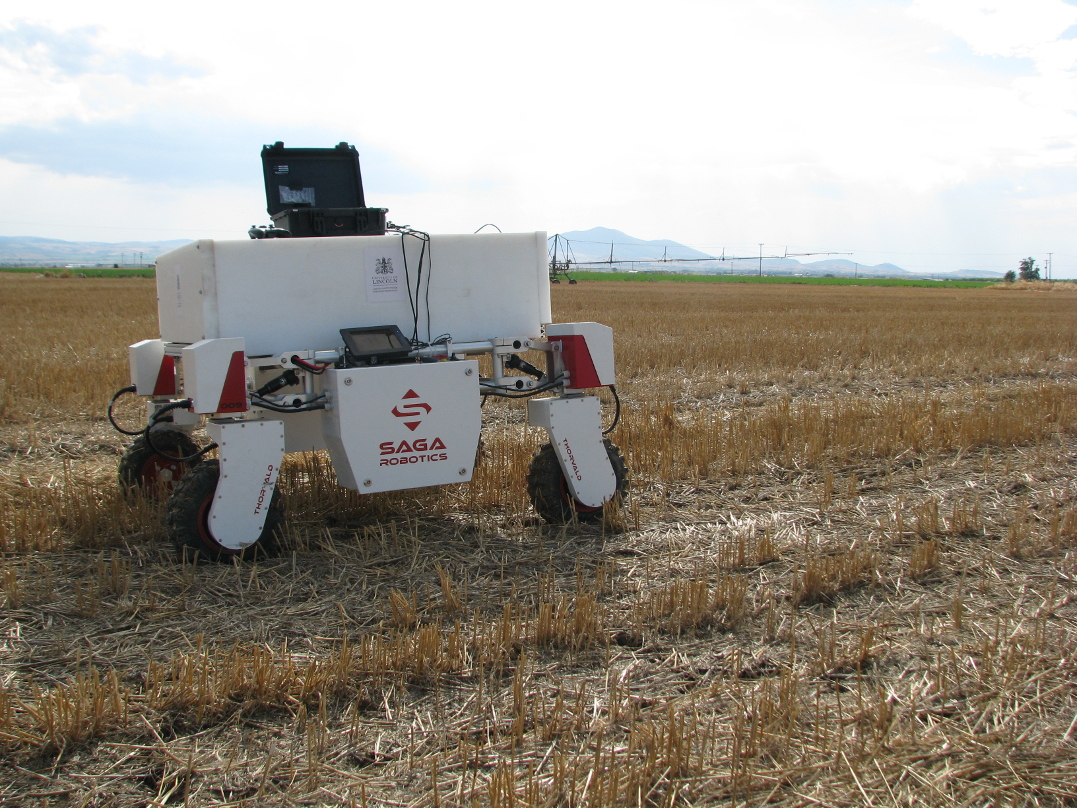}
      \caption{\label{fig:thorvald_greece}}
    \end{subfigure}
      \caption{The Thorvald robot equipped with a Cosmic-Ray sensor during data collection at: an airfield at the Lincolnshire Aviation Heritage Centre in East Kirkby, UK (left); a wheat stubble field near Volos, Greece (right).} \label{fig:thorvald}
\end{figure}

Our experimental set-up consists of an autonomous mobile robot Thorvald \cite{grmistad17thorvald} equipped with a custom-made, high sensitivity soil moisture sensor based on fast neutron counting principle manufactured by Hydroinnova (see Fig.~\ref{fig:thorvald}). The 12 neutron detectors are accompanied by temperature and humidity sensors which are used for providing the corrected neutron counts every 10 s. The sides and top of the sensor are shielded by using a 50 mm polyethylene shield to limit the detection footprint of the sensor to $~10$ m. The total weight of the sensor is around 300 kg. The sensor is interfaced with the robot through an Ethernet link. The robot is controlled through an in-built PC running Linux OS and Robot Operating System (ROS). The platform is equipped with a GNSS sensor, which enables robot localisation and geo-tagging of the collected data samples. The navigation component uses a graph-based representation, allowing the robot to move between a pre-determined set of waypoints.

\subsection{Datasets} \label{sec:datasets}

Evaluating the performance of robotic exploration strategies is inherently difficult and previous work in that domain often relies on simulated experiments (e.g.~\cite{Santos4DSim}). In our case, we propose to use the `surrogate' models of soil moisture, based on data collected from two real fields with the described equipment.  We used the collected data in off-line `simulations' to compare different exploration strategies and understand their overall performance. Simulations using a surrogate model are a useful tool to compare exploration methods~\cite{fentanes2011algorithm,Soil_compaction_kriging_2018}, providing the `ground truth' for the exploration results.

\begin{figure}[!ht]
	\centering
      \includegraphics[width=\columnwidth]{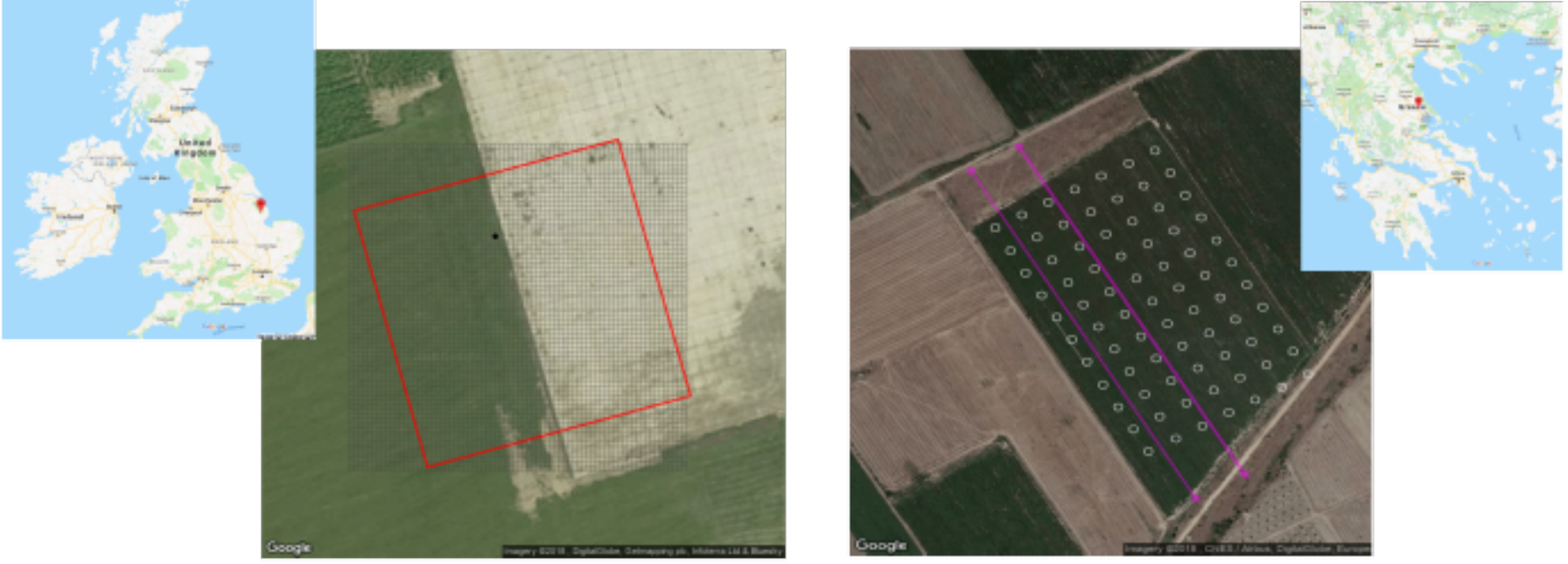}
      \caption{Location and layout of two data collection sites: an airfield (0.3 ha) at the Lincolnshire Aviation Heritage Centre in East Kirkby, UK (left); a wheat stubble field (7 ha) near Volos, Greece (right).}\label{fig:maps}
\end{figure}

The two data collection sites include an airfield at the Lincolnshire Aviation Heritage Centre in East Kirkby, UK and a wheat stubble field near Volos, Greece. Both fields were prepared in such a way so that they had equal parts of dry and wet land. Such an arrangement enabled us to systematically test the effectiveness of kriging-driven exploration strategies under a significant gradient between dry and wet areas akin to a step response.

The airfield site (see Fig.~\ref{fig:maps}) features a hard border between the grass field and concrete airstrip. Since concrete contains low levels of hydrogen, the airstrip provides a perfect replacement for dry conditions (5\% Volumetric Water Content (VWC)). The data collection took part in March 2018 and therefore the grass field was in a relatively wet condition (20\% VWC). 13 measurement locations where selected along a perpendicular line to the wet/dry border at 1, 2, 4, 8, 15, 30 m away from the border in both sides and a single point at the border itself. The measurement interval for all the points was set to 10 min.

The wheat stubble field in Greece (see Fig.~\ref{fig:maps}) covered a rectangular area of approx. 7 ha. The data collection took part in June 2018 under dry weather conditions. To create a wet area, the field was irrigated prior to data collection resulting in a wet/dry border with VWC of 18 \% for the dry part and 24 \% for the wet area, representing a fairly low gradient between the two parts. The whole field was meshed into a grid of 72 sampling locations with a spatial resolution of $30 \times 30$ m. The measurement interval for all the points was set to 10 min.

\begin{figure}[!ht]
	\centering
    \begin{subfigure}[b]{0.30\columnwidth}
      \includegraphics[height=0.25\textheight]{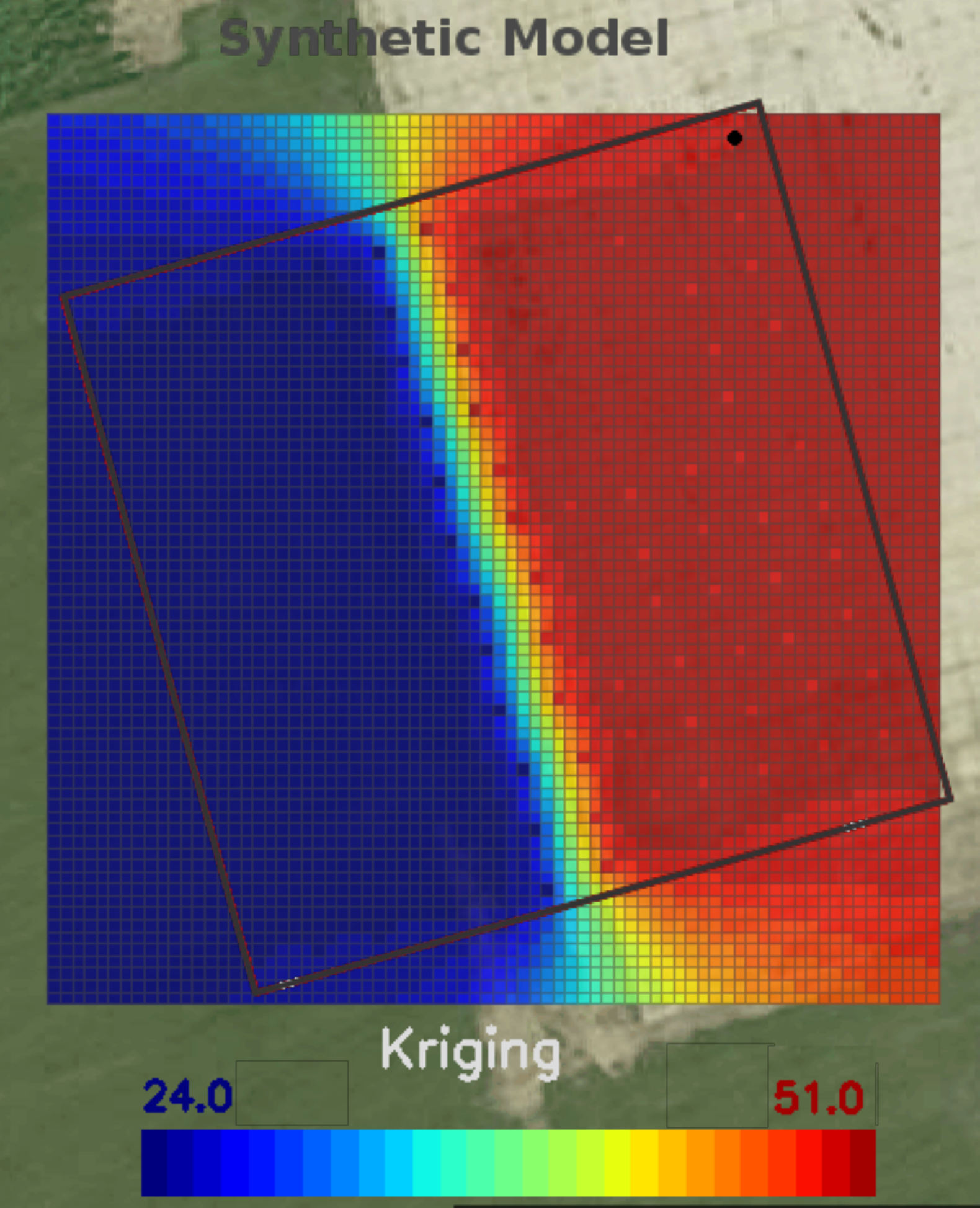}
      \caption{\label{fig:synthetic_ref}}
    \end{subfigure}
	~
    \begin{subfigure}[b]{0.30\columnwidth}
      \includegraphics[height=0.25\textheight]{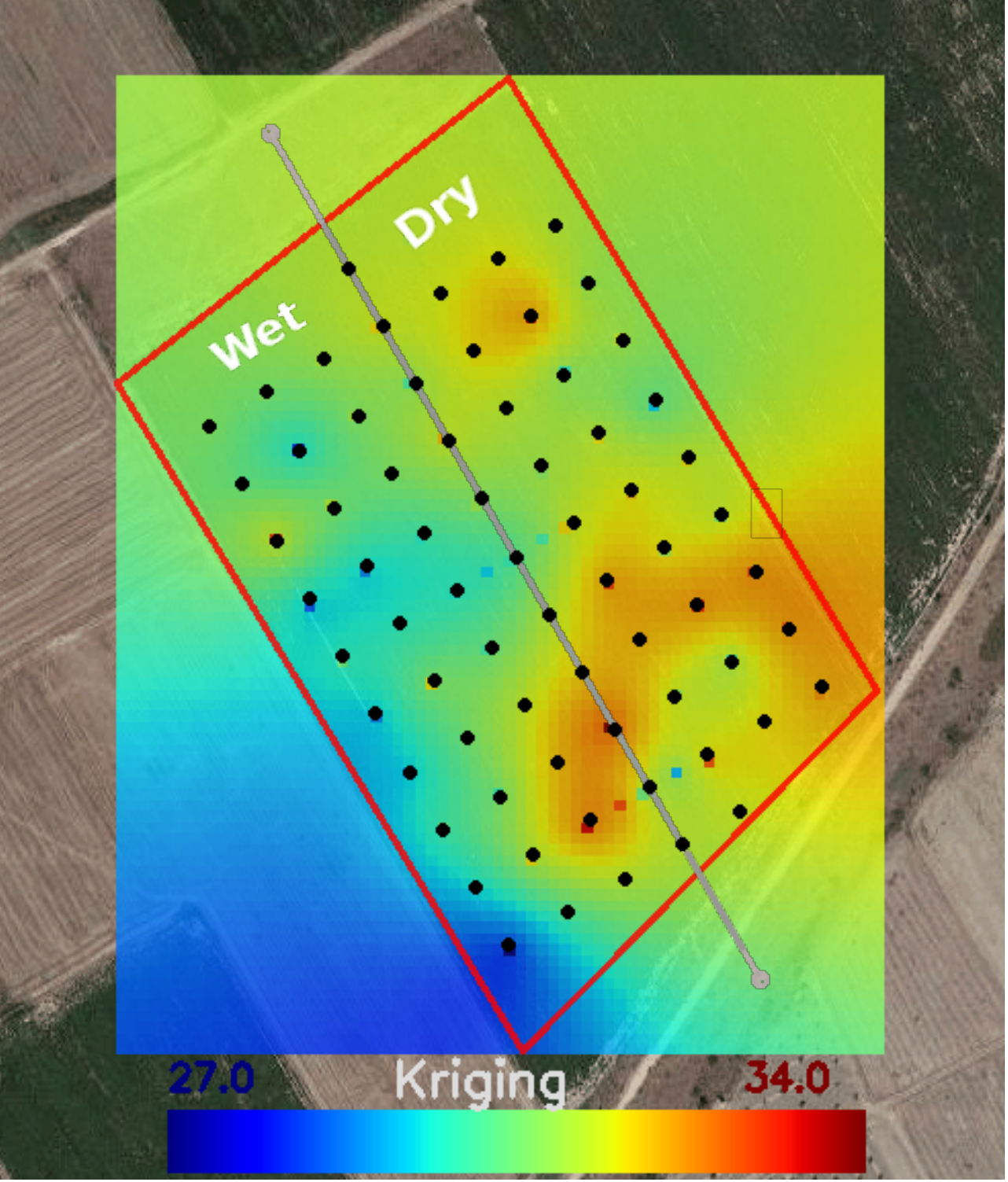}
      \caption{\label{fig:validation_ref}}
    \end{subfigure}
      \caption{The high gradient synthetic model generated from the airfield (a) and the validation model generated from the wheat stubble field (b).}\label{fig:real_model}
%
%
\end{figure}

Both datasets were used to create a set of testing models which were used to verify multiple hypotheses presented in Sec.~\ref{sec:results}. Each one of this testing models has neutron rates as inputs for the measurement model which were then extrapolated across the testing area using Ordinary Kriging (OK). This way an estimated rate can be produced for every location on the field. Once this is done the extrapolated rate is used as $\lambda$ to produce simulated readings every 10 s (real sensor's update rate) at specific locations using a Poisson distribution, resulting in a high density models used as a reference.
The models include:
\begin{itemize}
\item \emph{Synthetic model} which is based on real sensor rates recorded from the airfield (see Fig.~\ref{fig:synthetic_ref}). We generated two models representing a high and low gradient between sensor rates for wet and dry soil respectively. The high gradient corresponds to rates of 2.5 and 5.0 counts/s for the wet and dry part respectively. For the low gradient the values are 3.0 and 4.0 counts/s respectively. 

\item \emph{Simulated model} is based on the real data recorded in the airfield and extrapolated into multiple lines covering a rectangular area. The rates recorded along the single line crossing the wet/dry border are used to generate additional 5 parallel lines 10 m apart.

\item \emph{Validation model} in which the real data from the wheat stubble field is used (see Fig.~\ref{fig:validation_ref}). This model represents the most realistic soil moisture conditions and is used to validate the proposed algorithms.
\end{itemize}

\section{EXPERIMENTS} \label{sec:results}

To evaluate our framework, we have devised a set of experiments to test multiple hypotheses. First, that the robot will focus on sampling the area with the highest uncertainty, i.e. the border between the soil and concrete parts of the field and borders of the field. Second, we want to verify how much does the rate difference between the wet/dry parts of the field influence the exploration process (we call this a step response). Finally, we want to analyse the different impact of having a Fixed Measurement Interval (FMI) and an Adaptive Measurement Interval (AMI) which warrants a minimum measurement uncertainty before moving on to the next sampling point. Because our sensor follows the Poisson distribution model, we believe that the robot will require less time to sample the dry area of the field as it would have observed a higher number of events in the same time reducing the measurement $\sigma$.

The results presented in this section were obtained using simulated runs over the testing models presented in Section~\ref{sec:datasets}. The performance of the exploration methods presented in this section is evaluated in terms of travelled distance and model error. For assessing the quality of the resulting model, we compare the model produced against the surrogate model used for the exploration. To compare any two resulting models $A$ and $B$ we use Mean Square Error (MSE):
\begin{equation} \label{eq:rmse}
MSE=\frac{1}{n}\sum_{i=1}^{n}\left ( A_i - B_i\right )^2.
\end{equation}

\subsection{Fixed vs Adaptive Measurement Interval} \label{sec:fixVami}

To compare the influence that the sampling regime has over the exploration process, a greedy strategy was tested in the synthetic experimental set-up following four different sampling regimes, two Fixed Measurement interval (FMI) and two Adaptive Measurement Interval (AMI) experiments. For the FMI case one experiment was set to 10 minute intervals (FMI-long) and the other one to 5 minute intervals (FMI-short). For the AMI case one experiment was set to a $2.5\%$ measurement $\sigma$ threshold (AMI-long) and the other one to a $3\%$ threshold (AMI-short). Short and long cases should have comparable measurement times between them. 

\begin{figure}[!ht]
	\centering
	\begin{subfigure}[b]{0.48\columnwidth}
      \includegraphics[width=\columnwidth]{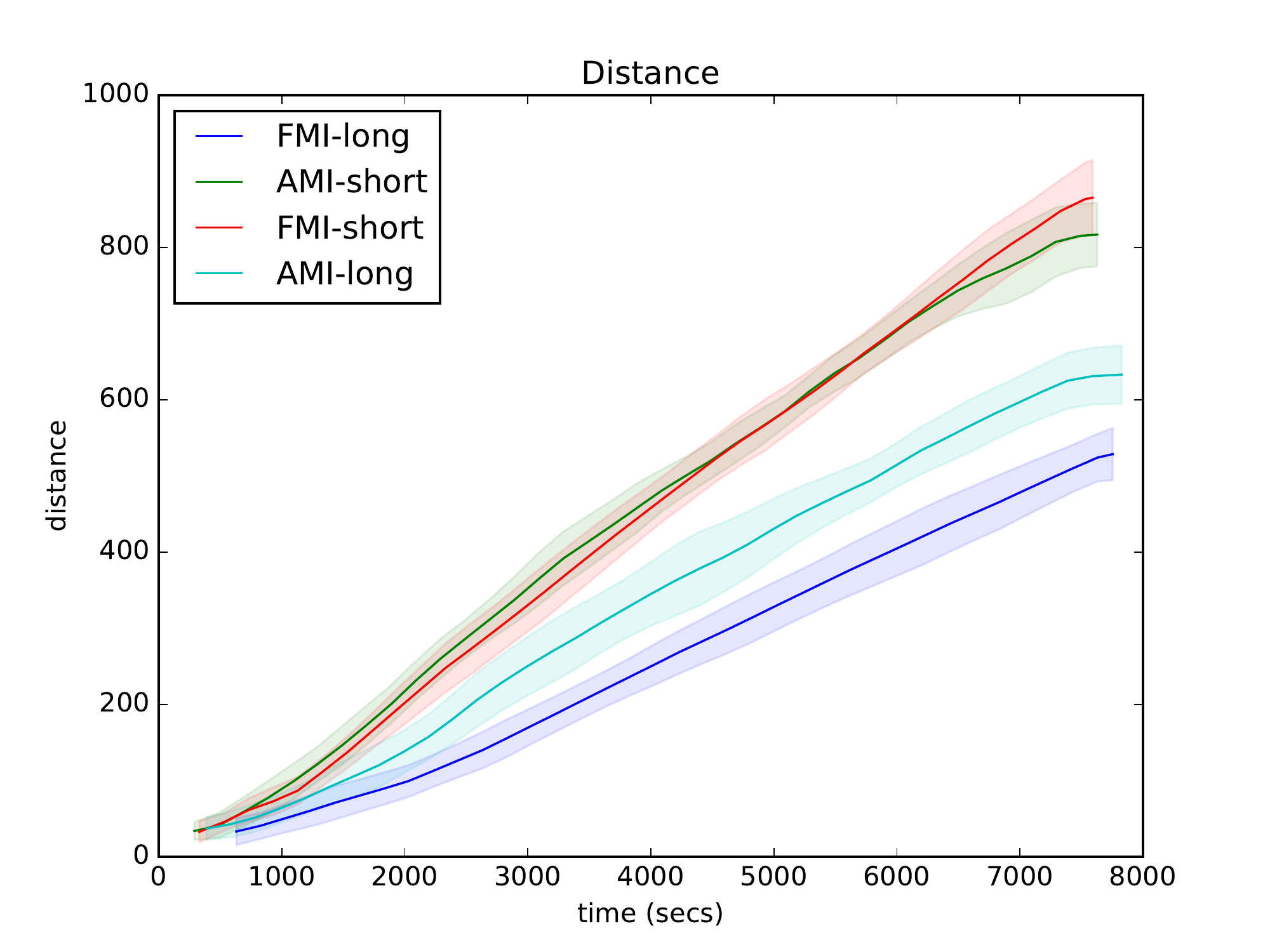}
      \caption{\label{fig:fix_ami_dist}}
    \end{subfigure}
	~
    \begin{subfigure}[b]{0.48\columnwidth}
      \includegraphics[width=\columnwidth]{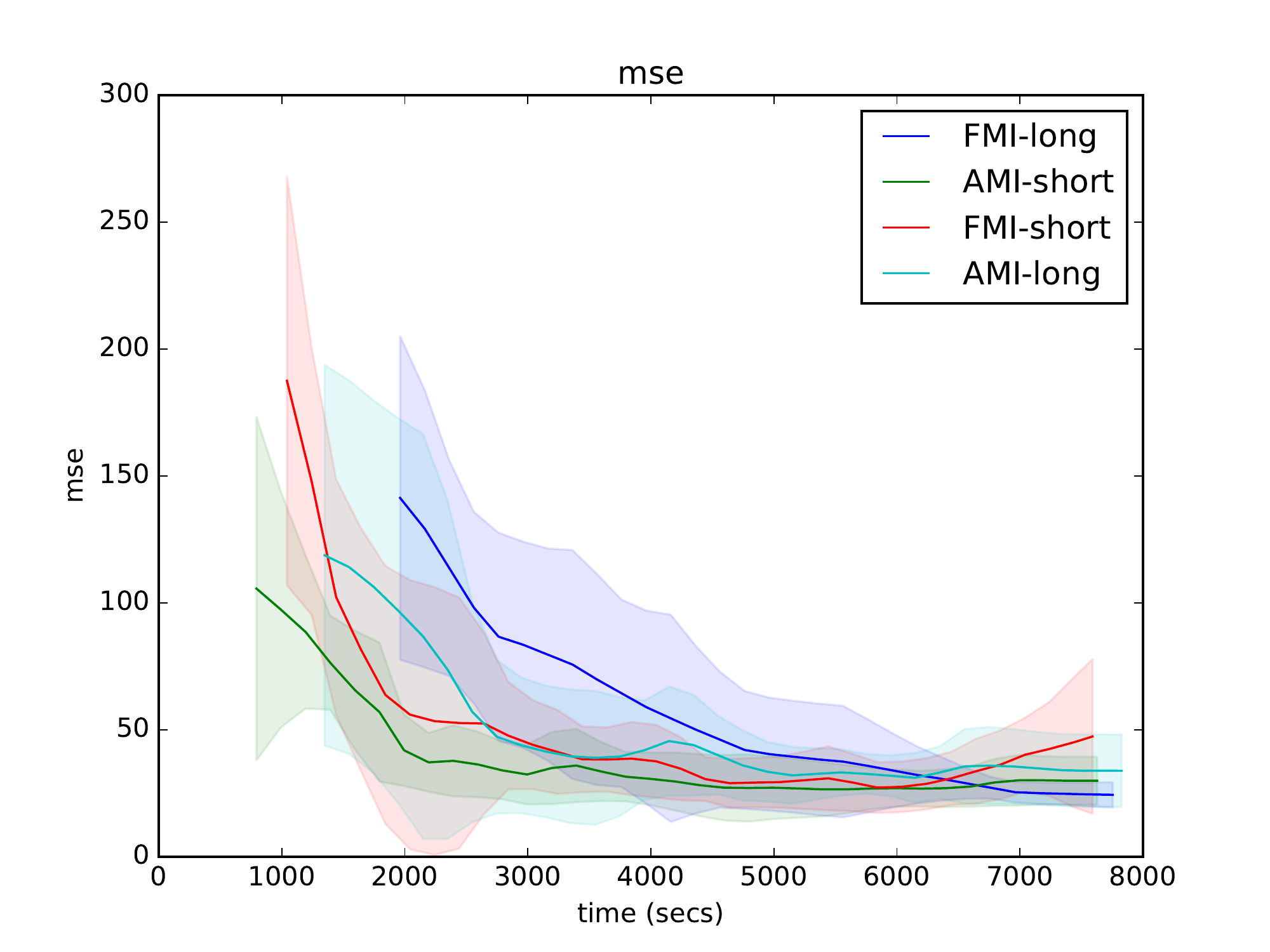}
      \caption{\label{fig:fix_asi_mse}}
    \end{subfigure}
    \caption{High-gradient synthetic scenario: comparison of performance for methods using Fixed vs Adaptive Measurement Interval in terms of (a) travel distance, and (b) Mean Square Error. Average results over ten runs, coloured areas represent standard deviation for each case.} \label{fig:fix_v_ami}
\end{figure}

Figure~\ref{fig:fix_ami_dist} shows that distance is driven mainly by measurement time.
This was predictable given that the amount of time that the robot spends collecting data is inversely proportional to the amount of time the robot spends navigating from one location to another. 
In figure~\ref{fig:fix_asi_mse} it can be seen that AMI regimes lead to faster convergence than their FMI counterparts.

Adaptive Measurement Interval strategies achieve better quality in shorter times because they can optimise the sampling time and drive exploration considering the conditions of the field (for example, the robot will spend less time in drier places as it will observe a higher number of events and achieve higher levels of confidence for the readings).
These gains are highly dependant on the variability of the soil moisture in the field, for example, in a predominantly wet field the gains from adaptive sampling interval strategies will be less noticeable. To verify this hypothesis, this analysis was also performed on a simulation with lower gradient between the wet and dry parts. 

\begin{figure}[!ht]
	\centering
	\begin{subfigure}[b]{0.48\columnwidth}
      \includegraphics[width=\columnwidth]{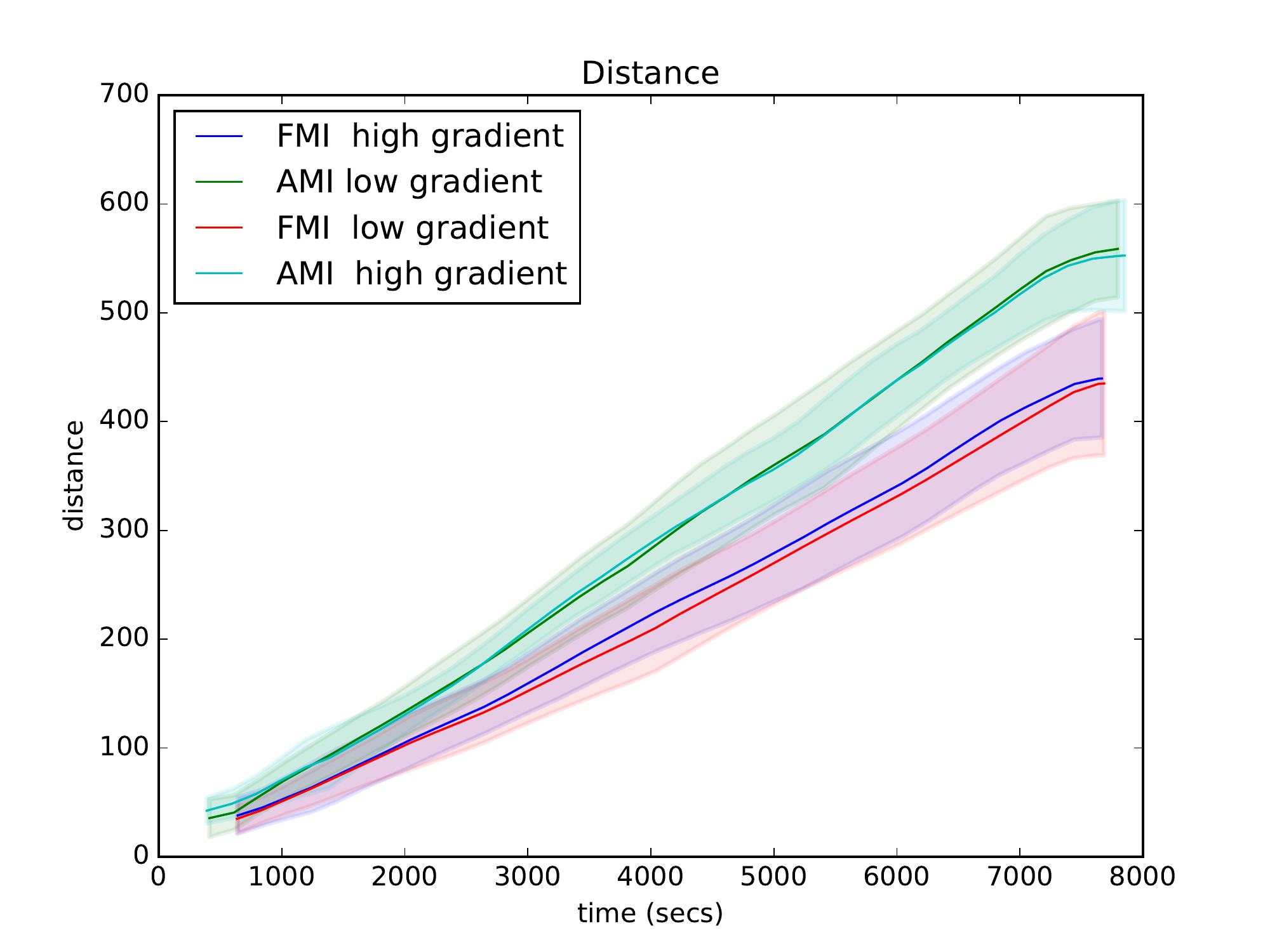}
      \caption{\label{fig:low_grad_dist}}
    \end{subfigure}
	~
    \begin{subfigure}[b]{0.48\columnwidth}
      \includegraphics[width=\columnwidth]{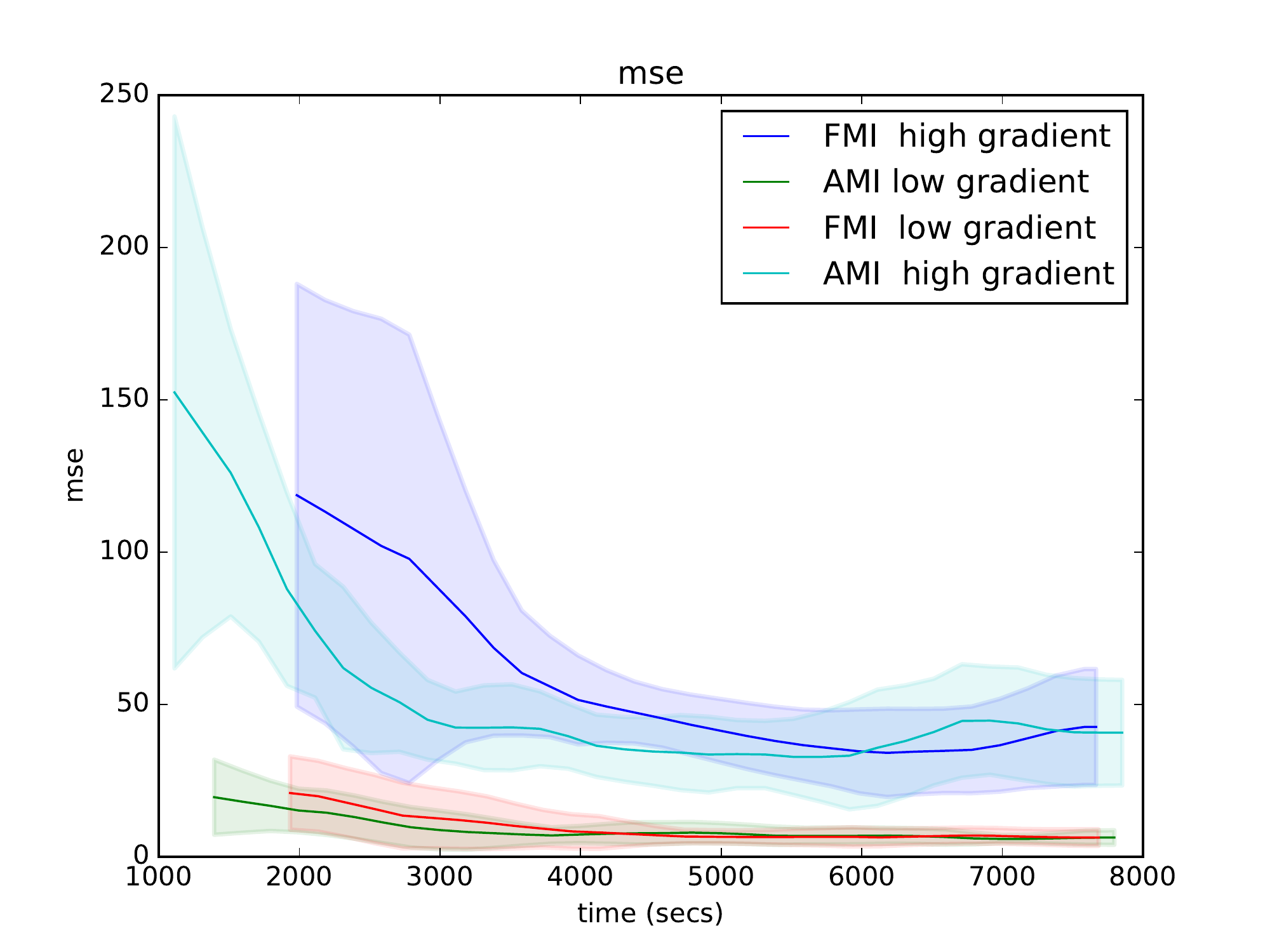}
      \caption{\label{fig:low_grad_mse}}
    \end{subfigure}
    \caption{Comparison of performance of Fixed and Adaptive Measurement Intervals on low- and high-gradient synthetic models in terms of (a) distance, and (b) Mean Square Error. Average results over ten runs, coloured areas represent standard deviation for each case.} \label{fig:fix_v_ami_low}
\end{figure}

Figure \ref{fig:fix_v_ami_low} shows a comparison of the performance of both regimes in scenarios with different gradients. From this figure it can be seen that the difference in performance between both regimes is not as noticeable in the scenario with the lower gradient. However, the travelled distance for the Adaptive regime is slightly higher in both cases, indicating that sampling regimes are not important for controlling the travelled distance, and that this is a factor that is probably driven by the exploration strategy.

\begin{figure}[!ht]
	\centering
	\begin{subfigure}[b]{0.48\columnwidth}
      \includegraphics[width=\columnwidth]{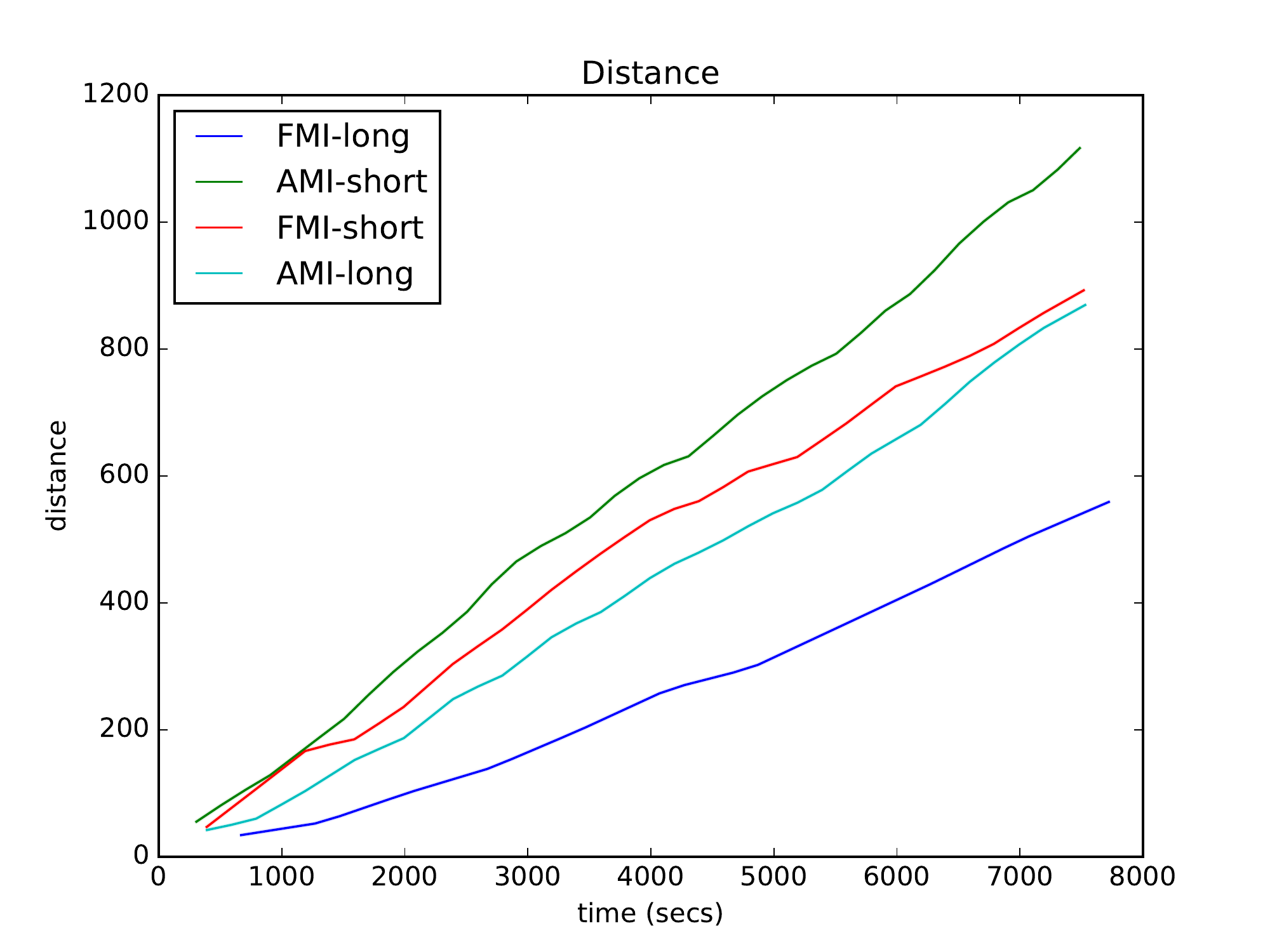}
      \caption{\label{fig:simu_fix_v_ami_dist}}
    \end{subfigure}
	~
    \begin{subfigure}[b]{0.48\columnwidth}
      \includegraphics[width=\columnwidth]{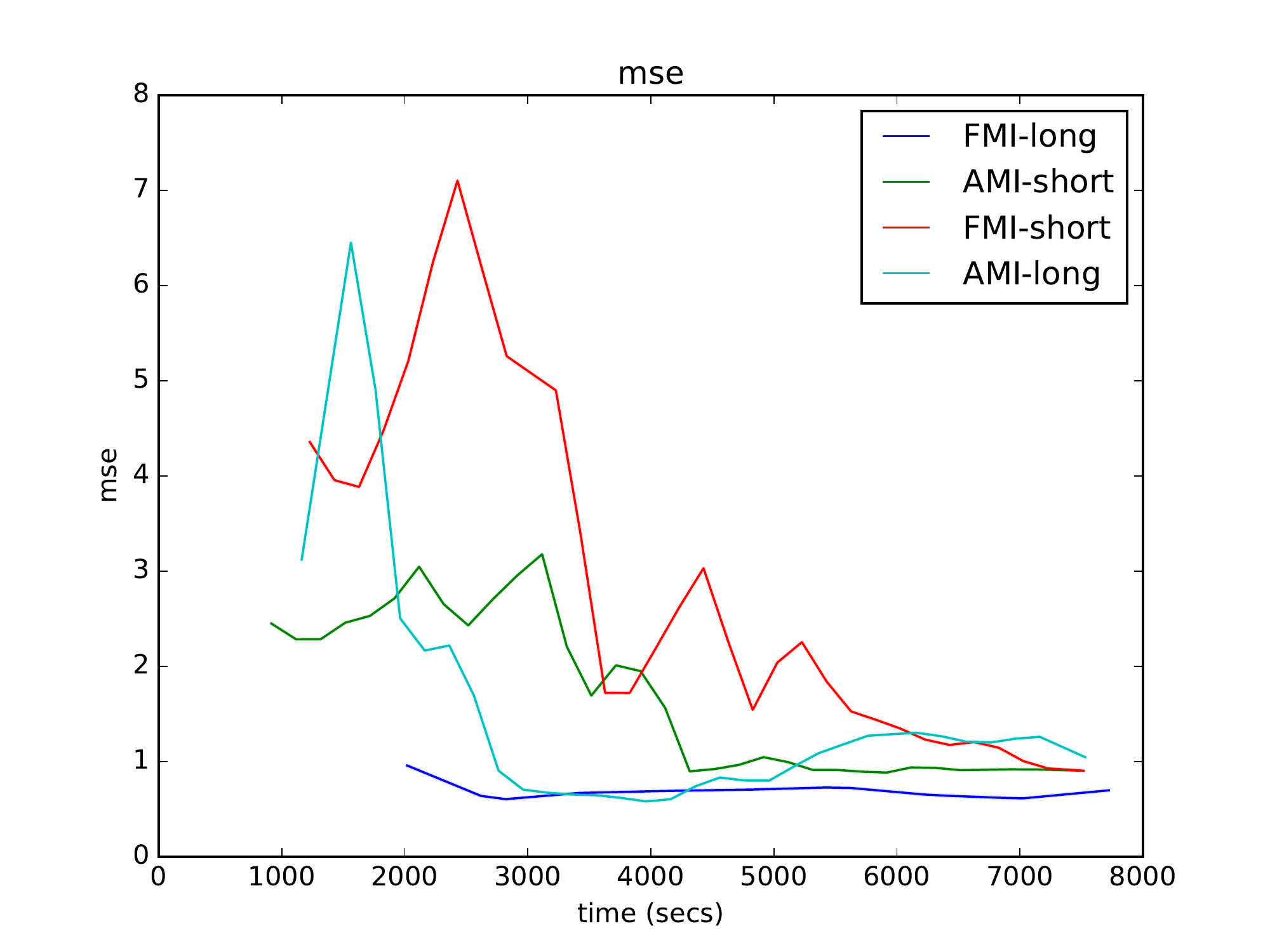}
      \caption{\label{fig:simu_fix_v_ami_mse}}
    \end{subfigure}
    \caption{Simulated scenario: comparison of performance for methods using Fixed vs Adaptive Measurement Interval in terms of (a) travel distance, and (b) Mean Square Error.} \label{fig:simu_fix_v_ami}
\end{figure}

Figure \ref{fig:simu_fix_v_ami}, presents a comparison between the performance of both sampling regimes in the simulated model. Comparing these results to the ones obtained with the synthetic model (Section \ref{sec:fixVami}), it is possible to see that the results are almost identical for both cases. This indicates that despite the fact that variability 
is just slightly higher than the lower gradient synthetic scenario, the sampling regime does have an influence, this could indicate that it is preferable to use an AMI regime in every case as it becomes influential with medium gradients but its cost in travel distance is not much higher.

\subsection{Comparison of the Exploration Strategies}

To verify the influence of different exploration strategies over the exploration process, we ran a series of simulations with 3 different strategies namely: Greedy, Monte Carlo and Adaptive Sampling. In all cases we used AMI as the measurement interval regime to isolate the effects of the exploration strategy only.

\begin{figure}[!ht]
	\centering
	\begin{subfigure}[b]{0.48\columnwidth}
      \includegraphics[width=\columnwidth]{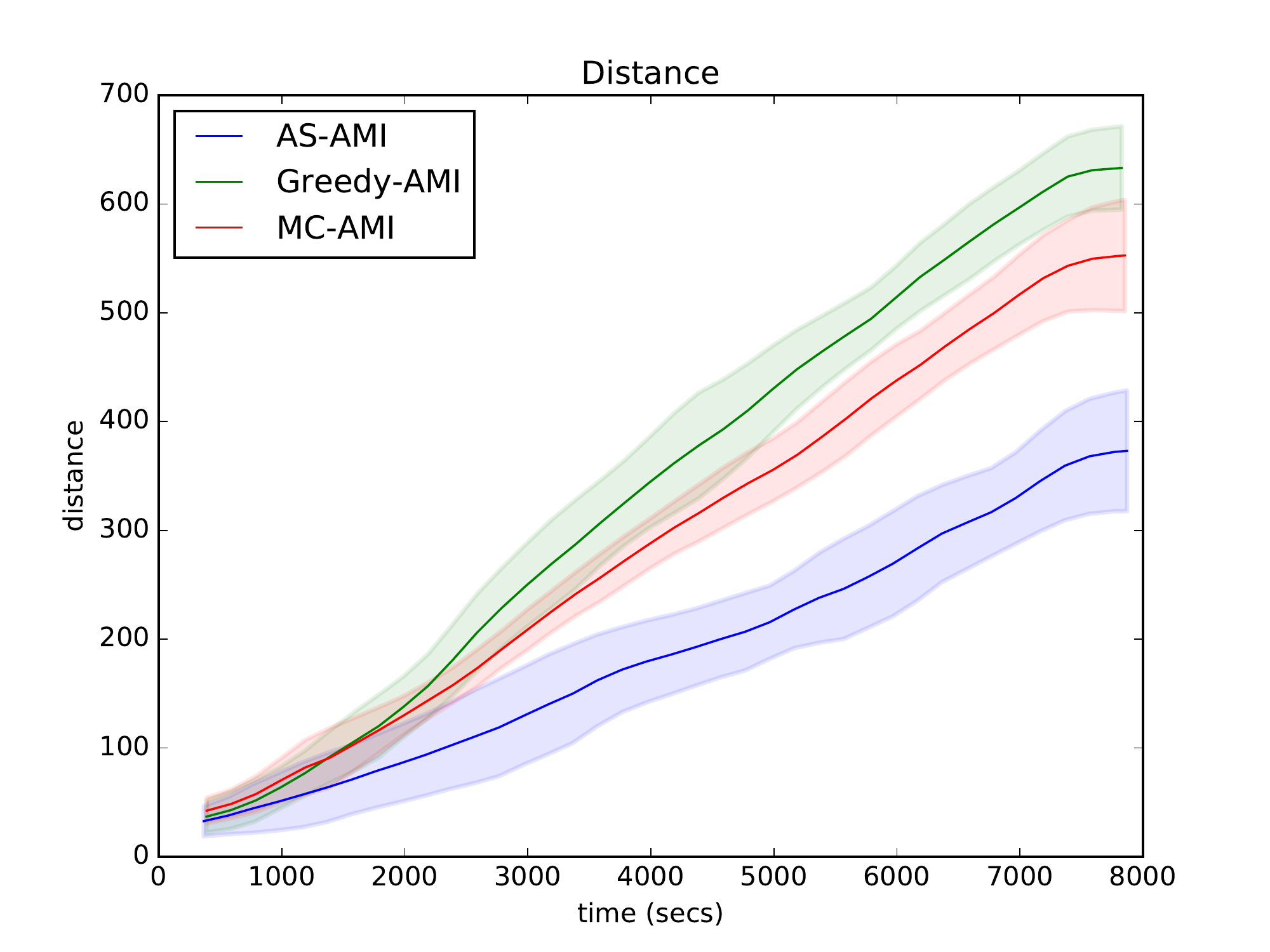}
      \caption{\label{fig:ami-strat_dist}}
    \end{subfigure}
	~
    \begin{subfigure}[b]{0.48\columnwidth}
      \includegraphics[width=\columnwidth]{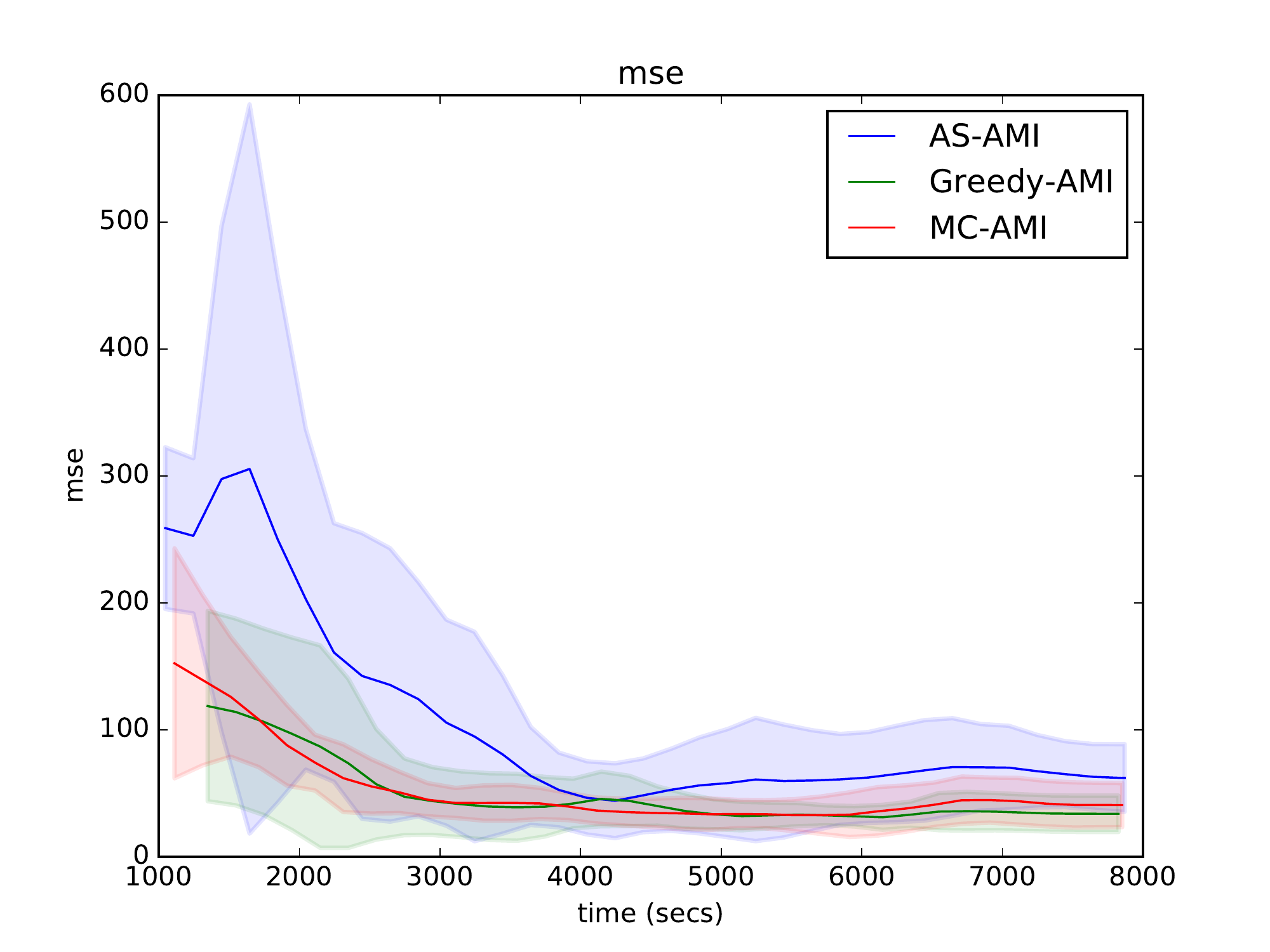}
      \caption{\label{fig:ami-strat_mse}}
    \end{subfigure}
    \caption{Synthetic airfield scenario: performance for different strategies using Adaptive Measurement Intervals in terms of (a) distance, and (b) Mean Square Error. Coloured areas represent standard deviation over ten runs.}\label{fig:str_comp}
\end{figure}

Figure \ref{fig:str_comp} shows the performance of the different exploration strategies. From these results, it can be seen that exploration strategies have a high influence on the distance travelled by the robot. In particular, it can be noticed that adaptive sampling strategies can achieve models that are just slightly worse than the other two strategies but travel much lower distances. This is a very important consideration, because in larger fields long travel distances can translate into significant amount of time not spent on gathering data, which at the end might end up with the lower quality models.

\begin{figure}[!ht]
	\centering
    \begin{subfigure}[b]{0.23\columnwidth}
      \includegraphics[width=\columnwidth]{figs/sythetic_out}
      \caption{\label{fig:sythetic_out}}
    \end{subfigure}
	~
    \begin{subfigure}[b]{0.23\columnwidth}
      \includegraphics[width=\columnwidth]{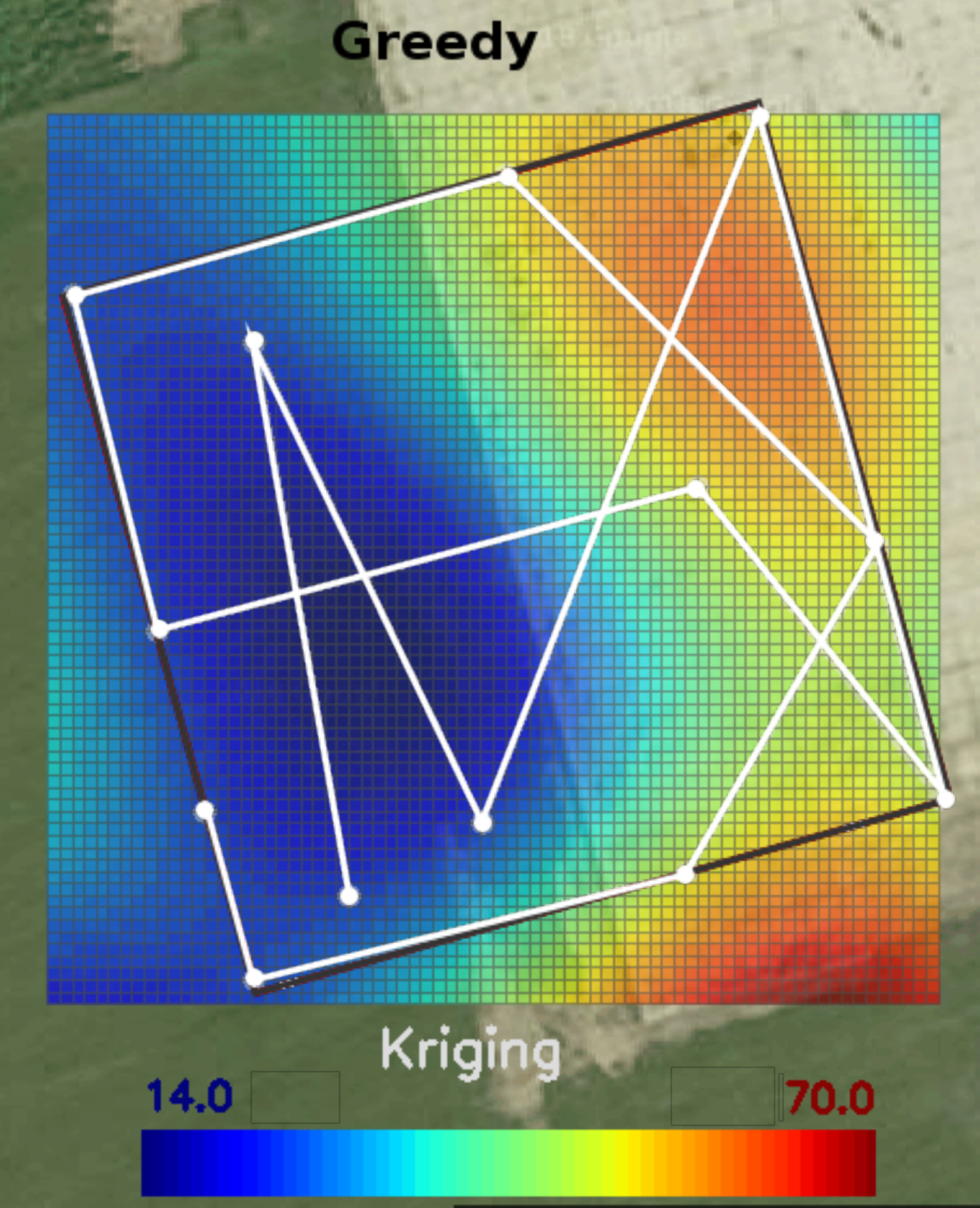}
      \caption{\label{fig:greedy_out}}
    \end{subfigure}
	~
    \begin{subfigure}[b]{0.23\columnwidth}
      \includegraphics[width=\columnwidth]{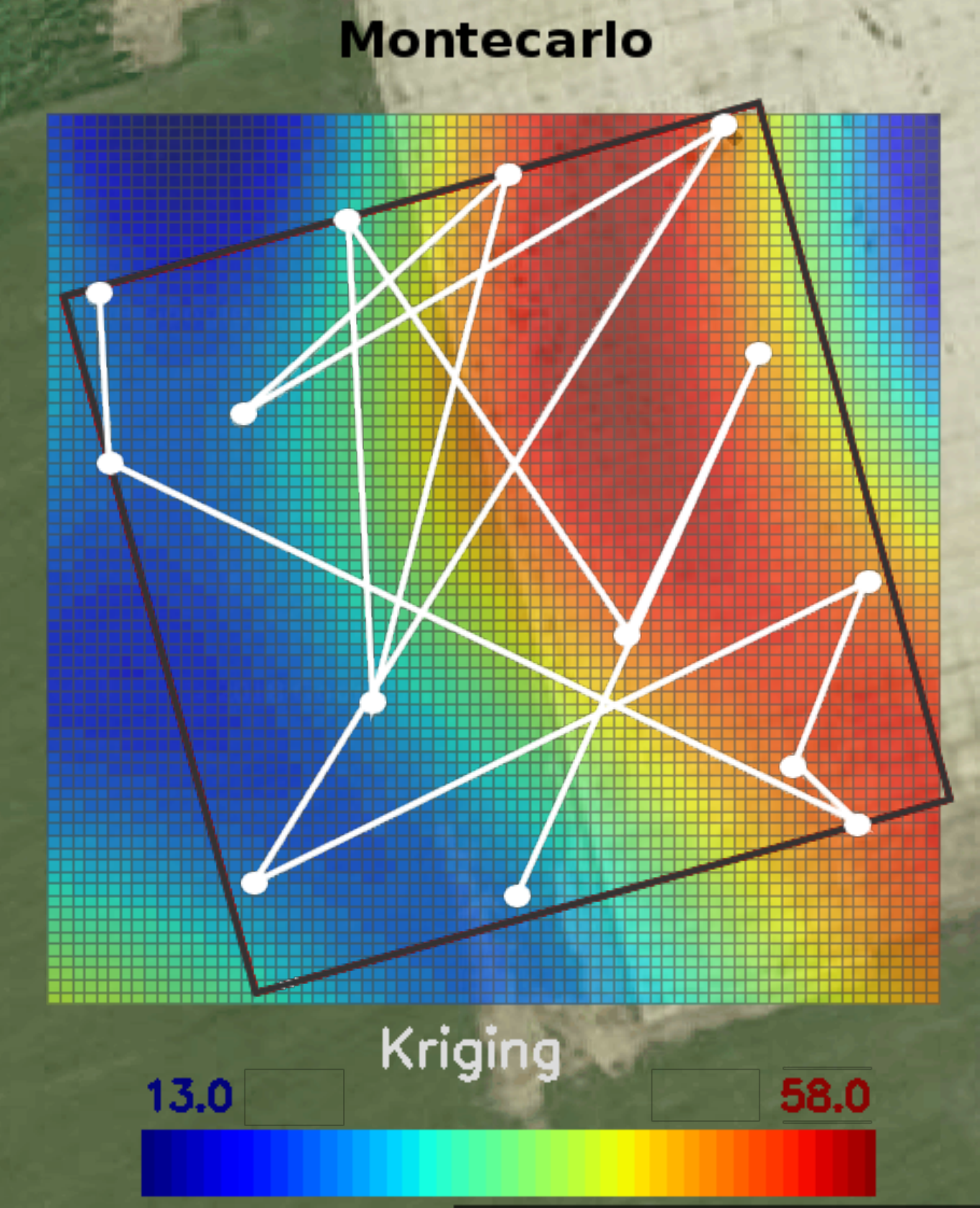}
      \caption{\label{fig:mc_out}}
    \end{subfigure}
	~
    \begin{subfigure}[b]{0.23\columnwidth}
      \includegraphics[width=\columnwidth]{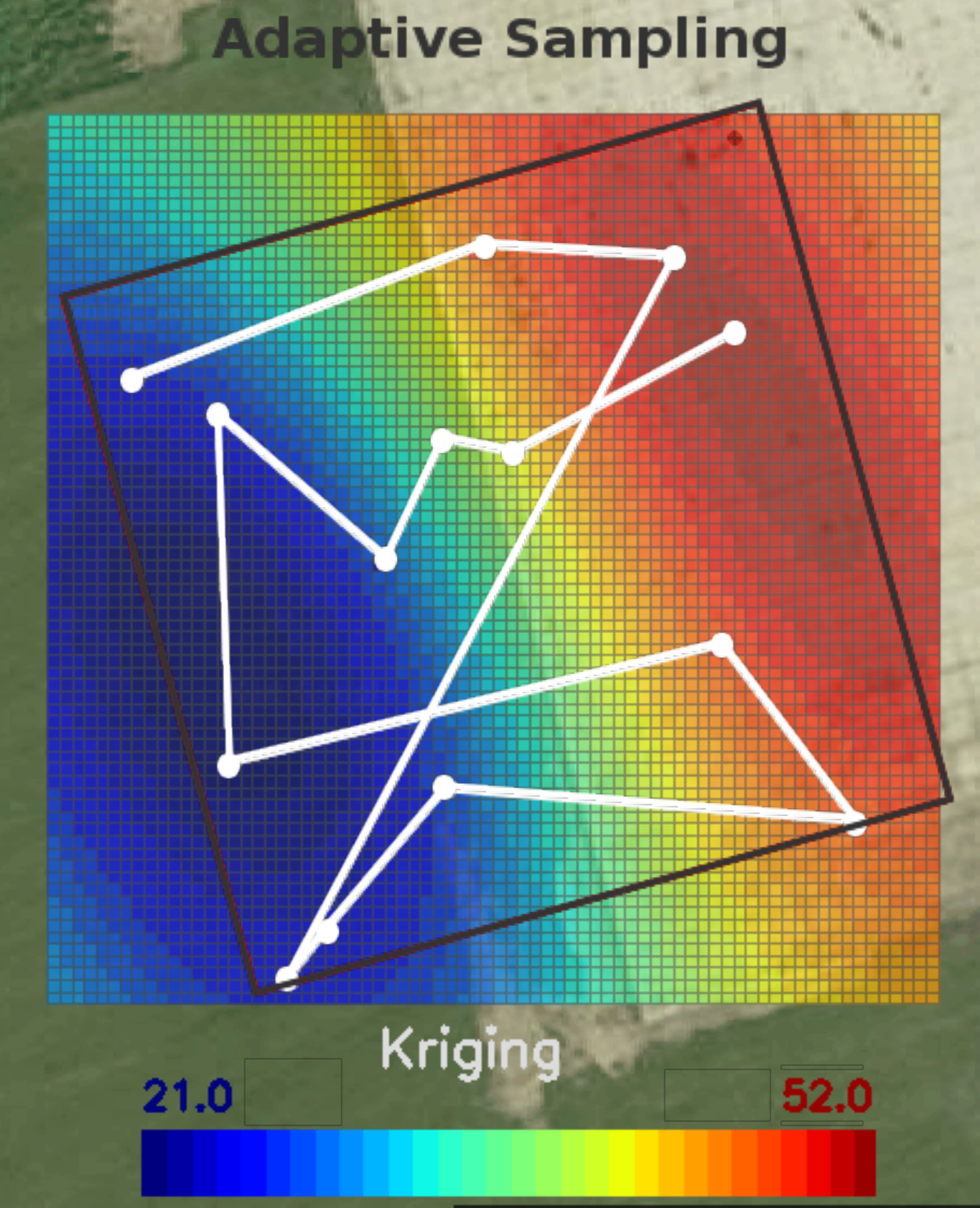}
      \caption{\label{fig:Adaptive_sampling_out}}
    \end{subfigure}
    ~
    \begin{subfigure}[b]{0.23\columnwidth}
      \includegraphics[width=\columnwidth]{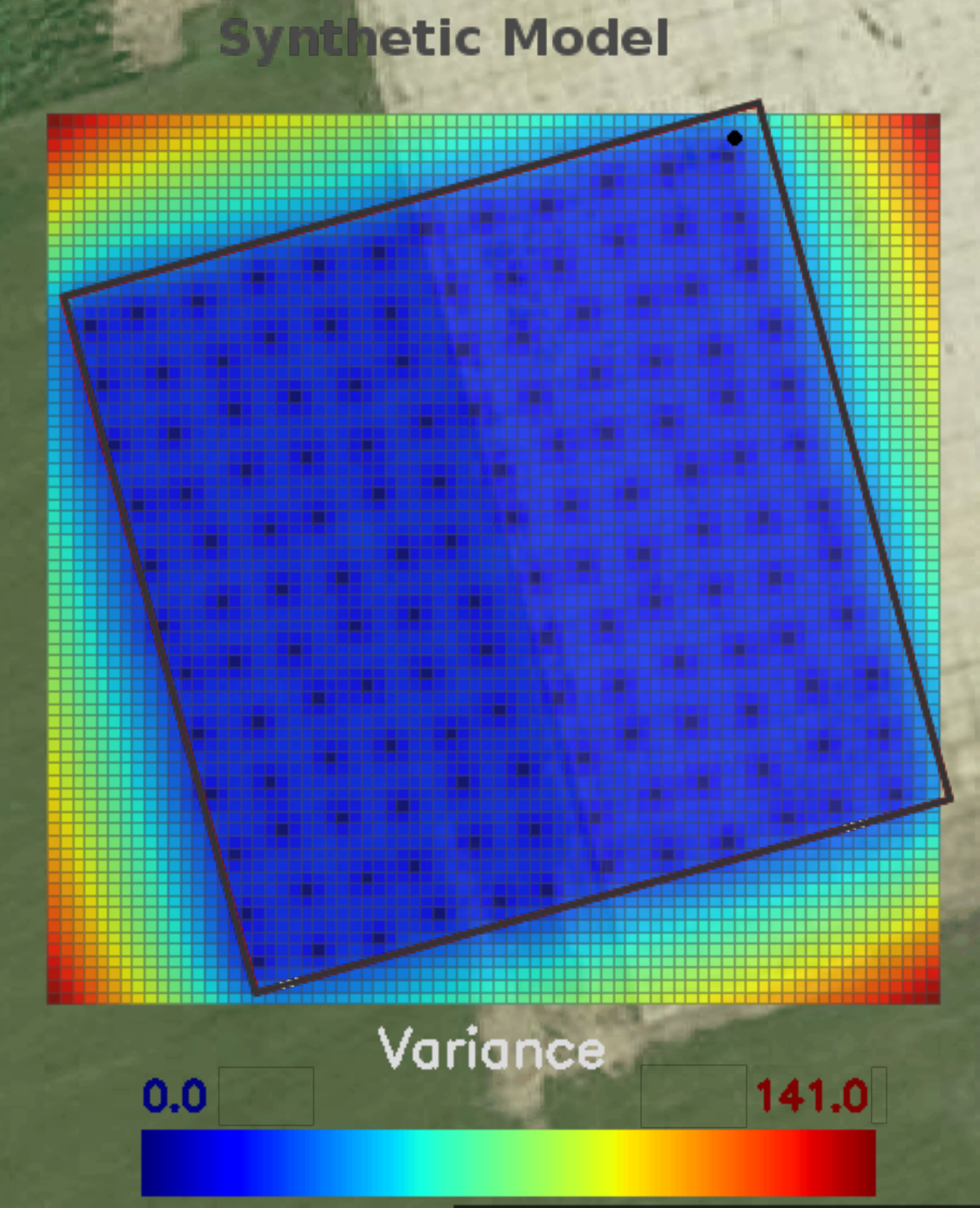}
      \caption{\label{fig:sythetic_var}}
    \end{subfigure}
	~
    \begin{subfigure}[b]{0.23\columnwidth}
      \includegraphics[width=\columnwidth]{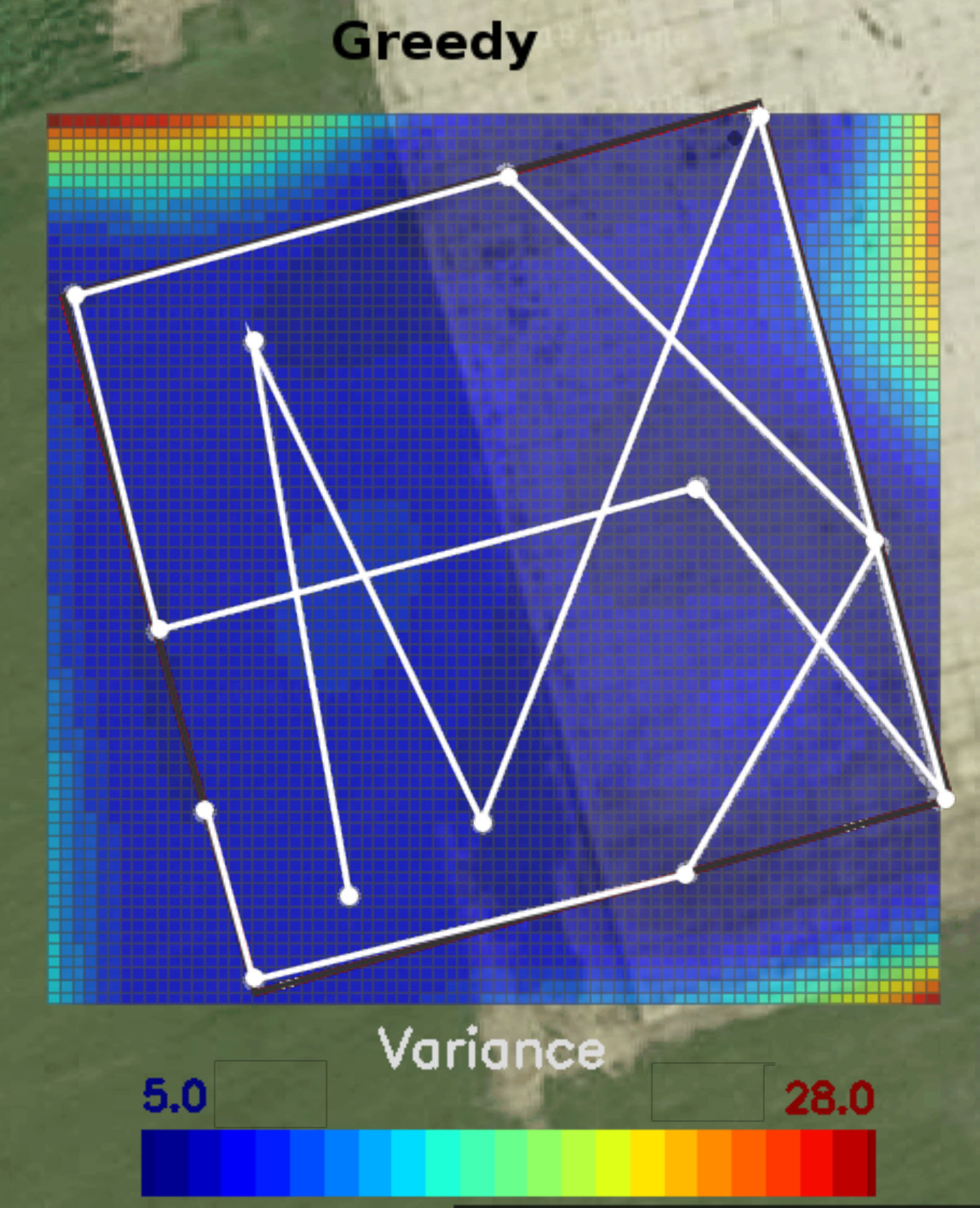}
      \caption{\label{fig:greedy_var}}
    \end{subfigure}
	~
    \begin{subfigure}[b]{0.23\columnwidth}
      \includegraphics[width=\columnwidth]{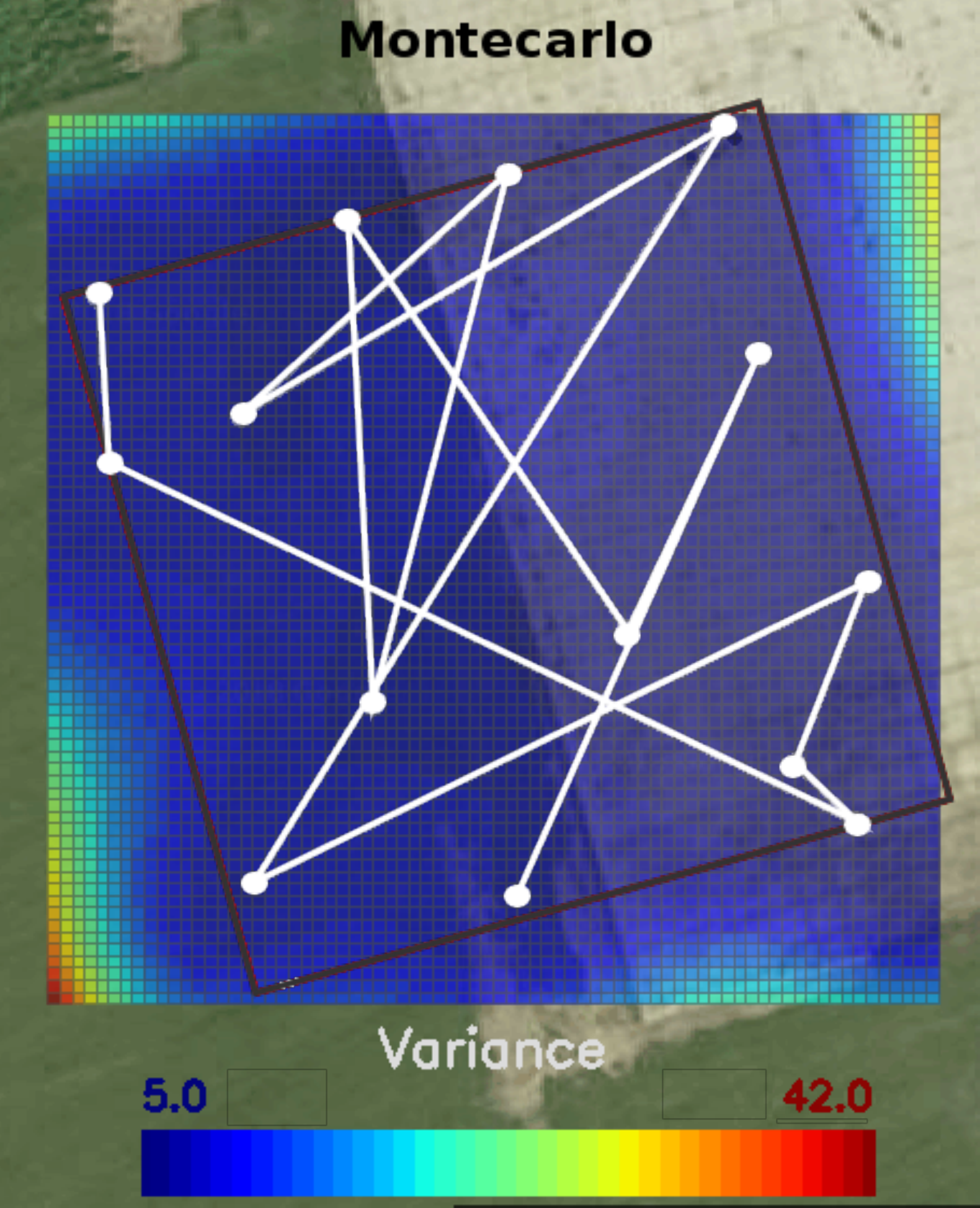}
      \caption{\label{fig:mc_var}}
    \end{subfigure}
	~
    \begin{subfigure}[b]{0.23\columnwidth}
      \includegraphics[width=\columnwidth]{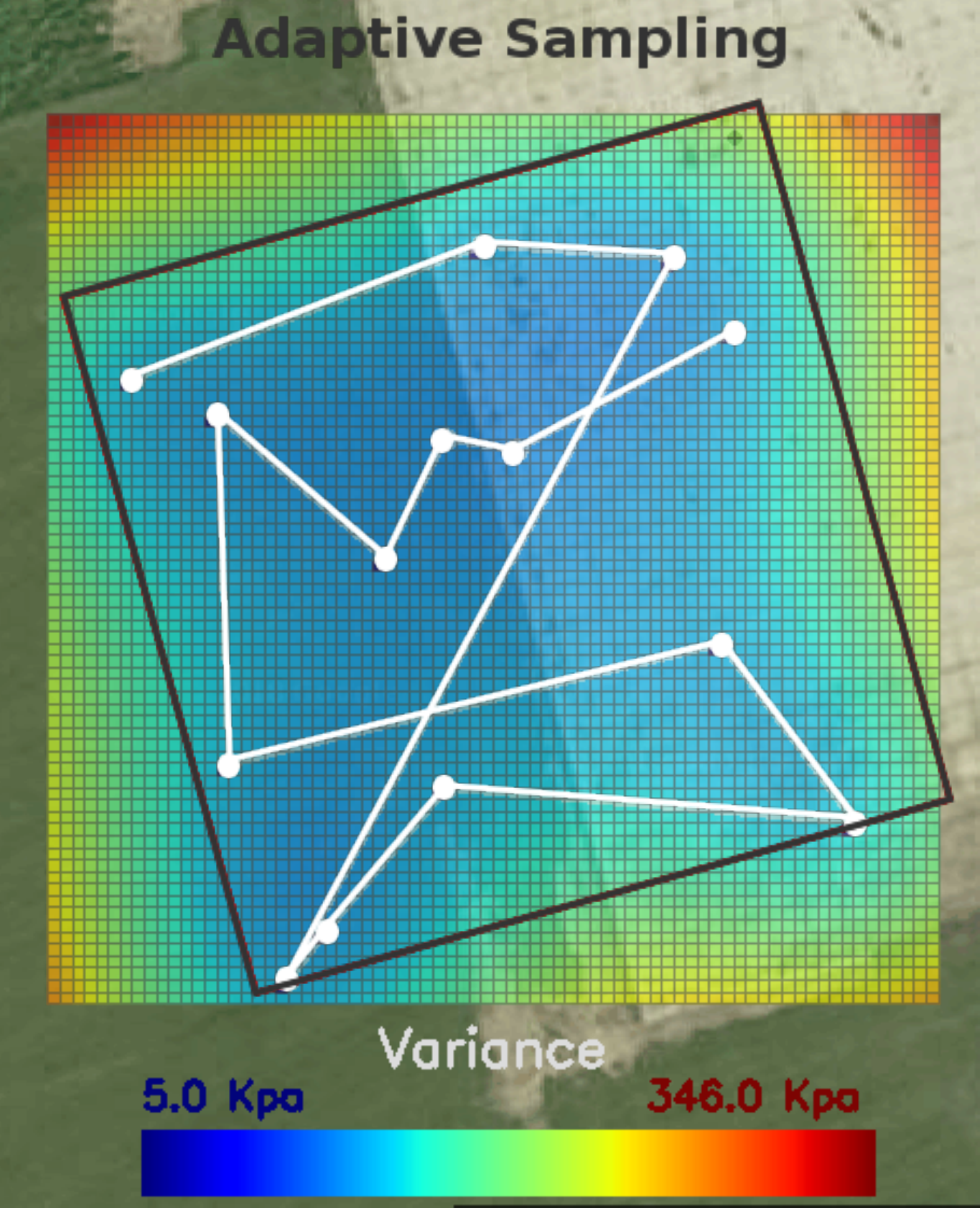}
      \caption{\label{fig:Adaptive_sampling_var}}
    \end{subfigure}
	\caption{High gradient synthetic model: exploration outputs and robot trajectories. The kriging output (top row) and variance (bottom row) for the full model (a/e), Greedy (b/f), Monte Carlo (c/g) and Adaptive (d/h) sampling strategies.} \label{fig:comp_outputs}
\end{figure}

Figure \ref{fig:comp_outputs} presents the outputs of the exploration result for the high-gradient synthetic model. The figure shows the resulting models for a field after two hours of autonomous exploration with the trajectories followed by the robot. One interesting thing is that the greedy strategy drives the robot mostly to the edges of the field. This is mainly because the kriging methods are better at interpolation than extrapolation, so the highest variances are always around the limit areas. This has the advantage that it can drive the model's variance down very quickly. It might also mean, however, that it can miss relevant infield information. In comparison, the adaptive sampling followed a much smoother and shorter trajectory and took samples that were better distributed across the field. 

\begin{figure}[!ht]
	\centering
	\begin{subfigure}[b]{0.48\columnwidth}
      \includegraphics[width=\columnwidth]{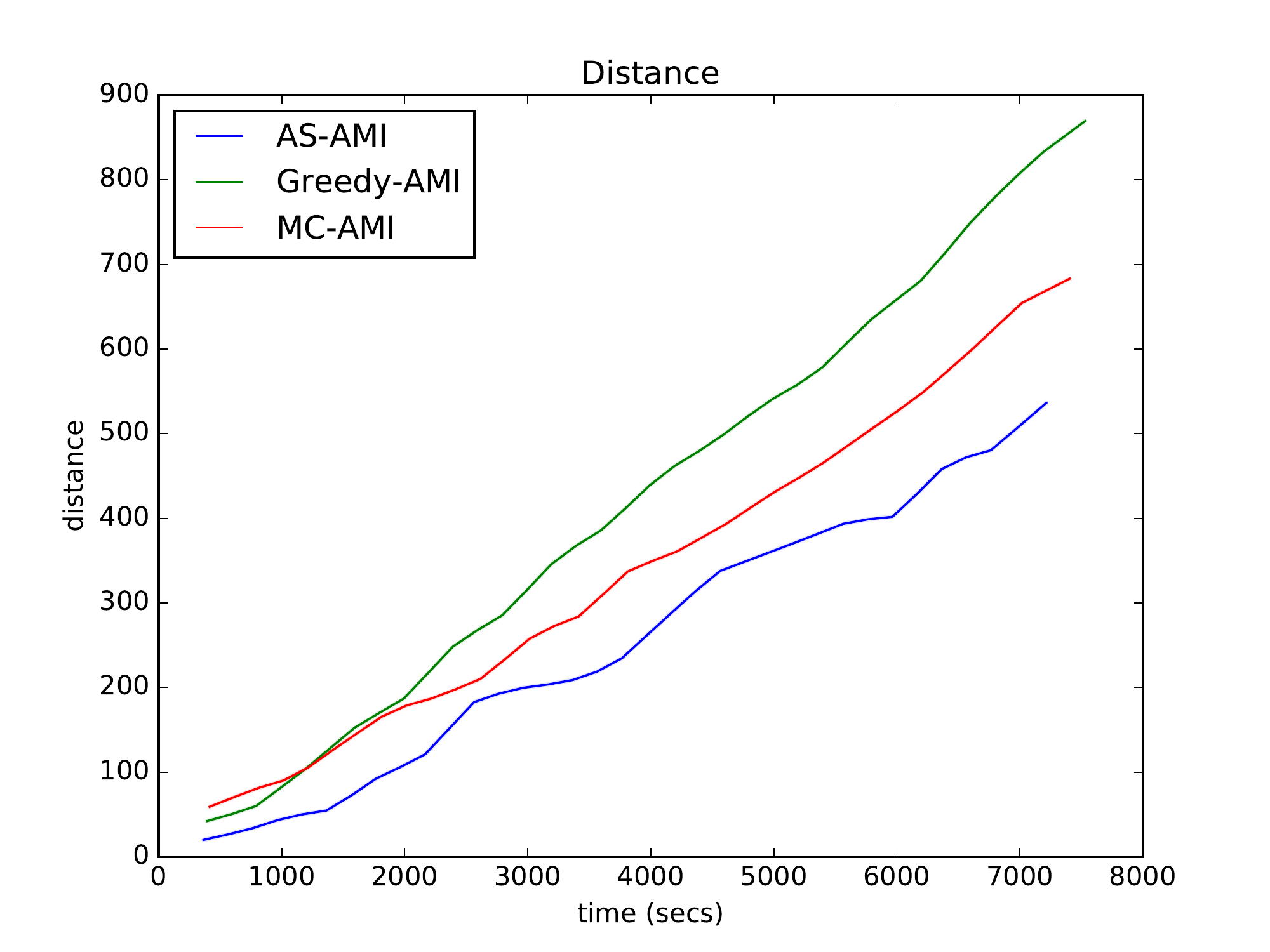}
      \caption{\label{fig:sim-strat_dist}}
    \end{subfigure}
	~
    \begin{subfigure}[b]{0.48\columnwidth}
      \includegraphics[width=\columnwidth]{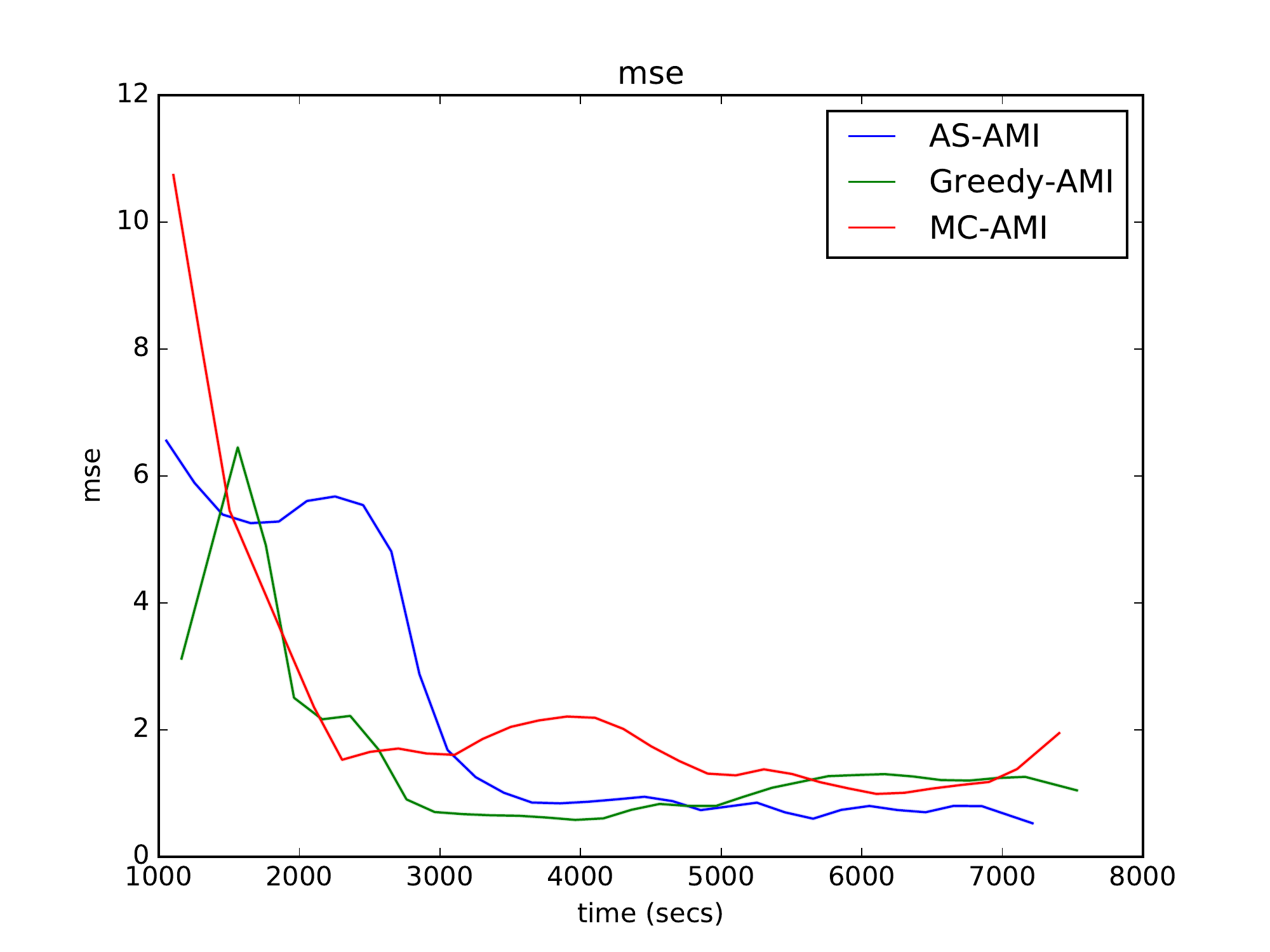}
      \caption{\label{fig:sim-strat_mse}}
    \end{subfigure}
    \caption{Simulated airfield scenario: performance for different strategies using Adaptive Measurement Intervals in terms of (a) distance, and (b) Mean Square Error. Coloured areas represent standard deviation over ten runs.}\label{fig:sim_str_comp}
%
%
\end{figure}

To verify this findings we performed the same test on the simulate airfield scenario. Figure \ref{fig:sim_str_comp} shows that the performance of the different strategies is similar to that exhibited in Figure \ref{fig:str_comp}. This indicates that the behaviour of each strategy is consistent and does not tend to vary much across testing scenarios.

\begin{figure}[!ht]
	\centering
      \includegraphics[width=0.9\columnwidth]{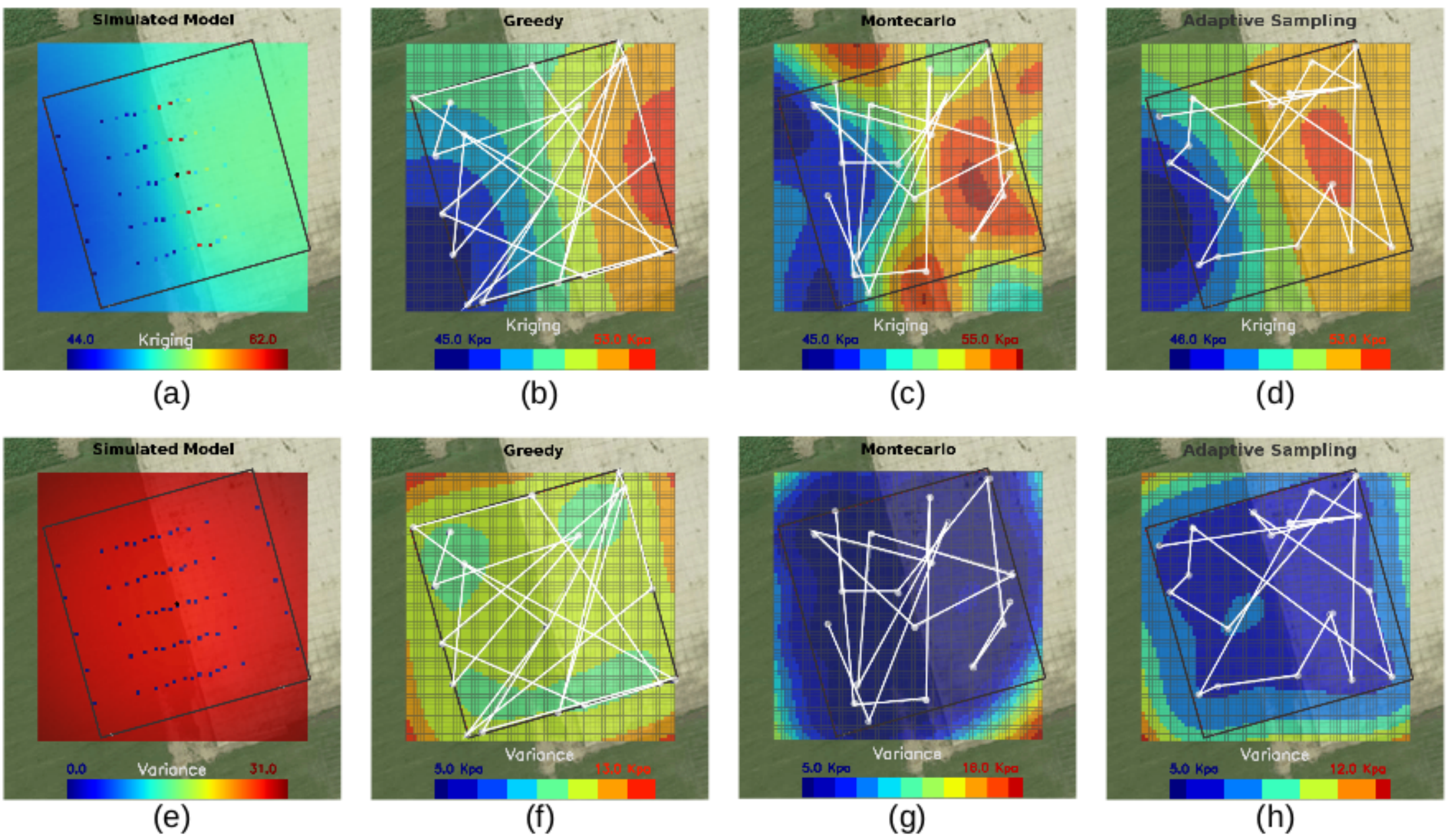}
      \caption{Simulated airfield model: exploration outputs and robot trajectories. The kriging output (top row) and variance (bottom row) for the full model (a/e), Greedy (b/f), Monte Carlo (c/g) and Adaptive (d/h) sampling strategies.}\label{fig:sim-maps}
\end{figure}

The outputs (see Figure \ref{fig:sim-maps}) show again that greedy strategies follow very long paths and outer sampling points contrasting to the adaptive sampling method which follows a more balanced approach, that seems to linger around areas that are either drier or wetter than usual.
The Monte Carlo approach shows an interesting behaviour,  it appears to be going back and forwards around the border between the grass and concrete, this seems to be because there higher variances around the border area however the paths are very random and the travelling distance is heavily penalised. In that sense the adaptive sampling strategy has a big advantage over Monte Carlo because it follows the same principle for choosing targets but at the same time it reduces travel distance.

\subsection{Validation on the Surrogate Model}

\begin{figure}[!ht]
	\centering
      \includegraphics[width=0.9\columnwidth]{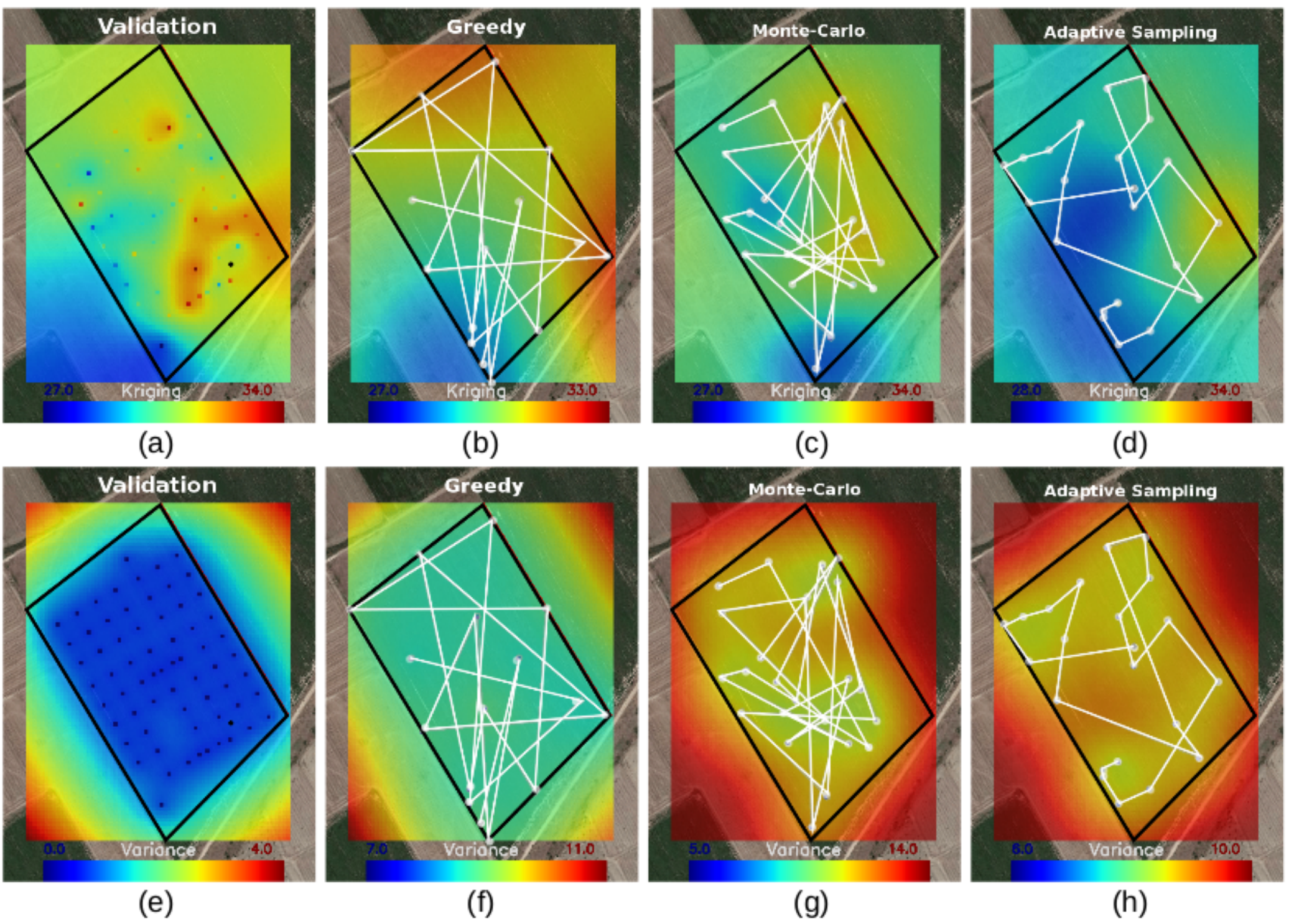}
      \caption{Validation model: exploration outputs and robot trajectories. The kriging output (top row) and variance (bottom row) for the full model (a/e), Greedy (b/f), Monte Carlo (c/g) and Adaptive (d/h) sampling strategies.}\label{fig:real-outs}
\end{figure}

To validate the methodology, several experiments were executed simulating an exploration task of four hours. Figure \ref{fig:real-outs} presents the resulting models for three experiments using different exploration strategies and AMI as sampling regime.

It is possible to see by simple visual inspection that the resulting models do not reflect perfectly the validation model. We believe that this is mainly due to two factors: first ,the gradient between wet and dry parts in this environment was very low which made it hard for the methods to identify areas of high variance. And second, the size of the environment limits how many samples per hectare the robot can achieve. This means in practice that the maps had much lower resolution than the validation model, hence each sample represents a much broader area.

This being said, it is worth noting that all the strategies generated models whose wetter areas and dryer areas correspond to those of the validation model. Also the soil moisture maps produced give a very good estimation of the areas were water deficit and concentration are in the field. Most likely, the miss-alignment between the validation model and the model outcome could have been overcome by having a longer mission. The fact that resulting model can discriminate wet and dry areas in such a short time (the validation model required more than 60 hours of work) is very encouraging.

\section{Conclusion}\label{sec:conclusions}
In this paper, we proposed an exploration framework for autonomous mobile robots equipped with a soil moisture sensor to create high quality soil moisture maps. The sensor is a novel device based on fast neutron counting which enables non-contact measurements of soil moisture. Such a class of sensors can be modelled by the Poisson distribution and we demonstrated how to integrate such measurements into the kriging framework. We also investigated a range of different exploration strategies and assessed their usefulness in different scenarios. The proposed framework was evaluated on a range of datasets based on real soil moisture data collected from two different fields.

One of the important findings of the paper is the fact that the sampling regime's contribution to the overall exploration process is highly dependant on the characteristics of the field. In fields with high variability and less uniform distribution of soil moisture, the use of Adaptive Measurement Interval shows significant improvements in model quality compared to a Fixed Measurement Time regime. We also demonstrated that adaptive sampling strategies guarantee lower navigation times and spend more time obtaining samples leading to models of comparative quality to the non-adaptive strategies but with a much lower travel distance. This is especially important in large fields where travelling takes a significant proportion of the exploration time. Greedy methods tend to sample the outer border of the environments, which is where the kriging variance is usually higher. They tend to miss localised patches, although their overall model quality is comparable. For small fields with uniform soil moisture distributions these might be preferable exploration strategies.

Although the presented framework was demonstrated for the soil moisture mapping, it is a general approach which can be used to map other soil properties such as compaction, chemical composure, etc. It is a framework that would be particularly suitable in scenarios where the measured phenomena directly affects the acquisition time and needs to be spatially mapped. This includes applications such as rainfall measurements, people and animal counting, gas detection etc. One of the follow up questions arising from this research is if changing the time measurement regime on the fly could improve the resulting models even further. Future work could also address the additional path planning constraints caused by the layout of typical agricultural fields which feature soil beds and rows. Finally, the framework will be extended to map multiple soil properties at the same time.





%

\section*{Acknowledgement}
This work was supported by the STFC Newton Fund programme, project ST/N006836/1.




%
\bibliographystyle{IEEEtran}
\bibliography{main}

\end{document}